\documentclass[pmlr]{jmlr}

\usepackage{longtable}
\usepackage{multirow}
\usepackage{graphicx}
\usepackage{caption}
\usepackage{subcaption}

 \usepackage{booktabs}

\usepackage[load-configurations=version-1]{siunitx} 

\makeatletter
\def\set@curr@file#1{\def\@curr@file{#1}} 
\makeatother

\theorembodyfont{\upshape}
\theoremheaderfont{\scshape}
\theorempostheader{:}
\theoremsep{\newline}

\jmlrvolume{xxx}
\jmlryear{2023}
\jmlrworkshop{}

\title[]{Optimizing Neural Network Scale for ECG Classification}

\author{
    \\
    \Name{Byeong Tak Lee}\thanks{Byeong Tak Lee and Yong-Yeon Jo contributed equally.}
    \Email{bytaklee@medicalai.com}\\ 
    \addr Medical AI Inc.\\
    Seoul, Republic of Korea 
    \AND
    \Name{Yong-Yeon Jo}\footnotemark[1]
    \Email{yy.jo@medicalai.com}\\ 
    \addr Medical AI Inc.\\
    Seoul, Republic of Korea
    \AND
    \Name{Joon-Myoung Kwon}
    \Email{cto@medicalai.com}\\ 
    \addr Medical AI Inc.\\
    Seoul, Republic of Korea
}

\begin{document}

\maketitle

\begin{abstract}

We study scaling convolutional neural networks (CNNs), specifically targeting Residual neural networks (ResNet), for analyzing electrocardiograms (ECGs). Although ECG signals are time-series data, CNN-based models have been shown to outperform other neural networks with different architectures in ECG analysis. However, most previous studies in ECG analysis have overlooked the importance of network scaling optimization, which significantly improves performance. We explored and demonstrated an efficient approach to scale ResNet by examining the effects of crucial parameters, including layer depth, the number of channels, and the convolution kernel size. Through extensive experiments, we found that a shallower network, a larger number of channels, and smaller kernel sizes result in better performance for ECG classifications. The optimal network scale might differ depending on the target task, but our findings provide insight into obtaining more efficient and accurate models with fewer computing resources or less time. In practice, we demonstrate that a narrower search space based on our findings leads to higher performance.
\end{abstract}

\section{Introduction}

Electrocardiograms (ECGs) are non-invasive diagnostic tests that record the electrical activity of the heart over time. They are used to detect and analyze various heart-related conditions. Over the years, several machine learning models have been developed for ECG analysis, including recurrent neural networks (RNNs) (\cite{sevilmis2017classification}), Transformers (\cite{yan2020ecg}), and convolutional neural networks (CNNs) (\cite{rajpurkar2017cardiologist}). Although ECG signals are considered multivariate time-series data (Figure \ref{fig:ecg}), recent studies predominantly employ CNN-based models instead of RNN-based models (\cite{sun2023towards, vaid2023multi}) breaking the convention.

\begin{figure*}[h]
    \centering
    \label{fig:ecg}
    \includegraphics[width=1.0\textwidth]{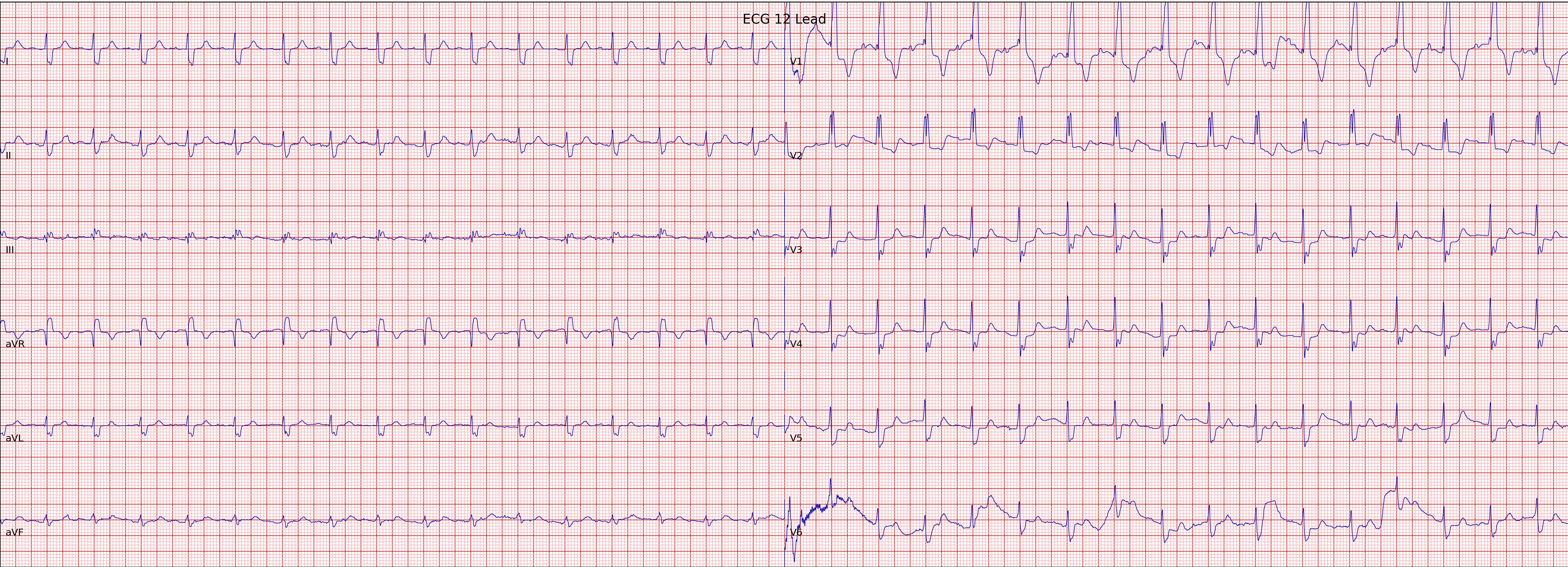}
    
    \caption{
    Example of 12-lead ECG signal.
    }
    \label{fig:ecg}
\end{figure*}

By adopting such models, they have achieved higher performance in ECG classifications (\cite{zheng2020optimal, elias2022deep, sun2023towards}). Despite their success, most previous studies have overlooked the importance of optimization in scaling networks. \textit{Optimizing network scale} refers to the process of tuning the architecture of a neural network by adjusting its scaling parameters. This process is critical for developing efficient and accurate models (\cite{koutini2021receptive}). In ECG classification, there is a single previous study evaluating the effect of scaling parameters of neural networks. It solely focuses on one aspect (e.g., the depth of the network) of network architecture (\cite{nonaka2021depth}).

To identify the importance of network scaling across various aspects, we examined the hyperparameters affecting network scale in ECG analysis. Before delving into the significance of scaling parameters, we first searched for widely used neural networks for ECG classifications. As a result, the residual neural network (ResNet) (\cite{tai2017image}) was chosen as a fundamental architecture because it is a typical model (\cite{rajpurkar2017cardiologist, nonaka2021depth, elias2022deep}) used in ECG classifications.

It is widely acknowledged that the depth of network layers, the convolution kernel size, and the number of input/output channels are the primary scaling parameters that influence ResNet's scale and complexity.
We conducted experiments to see how varying combinations of such parameters impacted the performance of ECG classification models on the Physionet2021 (\cite{reyna2021will}) and Alibaba (\cite{Alibaba}) datasets.

Our experiments show that models with a shallower network, a larger number of channels, and smaller convolution kernel sizes improve performance. This finding is significant because the performance changes depending on scaling parameters in ECG classification differ from those observed in image classifications~(\cite{zagoruyko2016wide}). Furthermore, optimizing network scale can be both time-consuming and require considerable computing resources. However, to attain better performance within a given time, configuring a narrow search space for scaling parameters based on our findings helps models approach the best possible outcome more efficiently. Additionally, we examined the performance variation caused by the optimization with respect to the labels of the dataset and investigated the reasons behind our findings.


\subsection*{Generalizable Insights about Machine Learning in the Context of Healthcare}

Determining the appropriate network architecture scale is critical for classification tasks. However, there hasn't been extensive research on evaluating network architecture scaling for analyzing ECGs. In our paper, we present a comprehensive experimental assessment of the performance of different network architecture scaling options for ECG classifications. Our findings reveal that models with a shallower network, a larger number of channels, and a smaller kernel size improve performance in ECG classifications. Interestingly, this finding differs from the existing knowledge in the computer vision domain. We claim that this could be a new guideline for efficiently tuning network scaling parameters in ECG analysis. Furthermore, we show that applying our findings to the hyperparameter optimization process improves the performance of ECG classifications. These results provide insights into practical architecture scaling in neural networks. We hope that these observations will assist clinical and engineering researchers in utilizing deep learning for ECG classification.

\section{Problem Definition}
\label{sec:format}
Although the optimization of network scaling is crucial due to its significant impact on model performance, existing ECG studies have overlooked this aspect (Appendix ~\ref{append:review}). Since ResNet (\cite{tai2017image}) is widely used for ECG analysis as a deep learning model (\cite{rajpurkar2017cardiologist, zheng2020optimal, nonaka2021depth, elias2022deep, sun2023towards}), we focused on scaling parameters for ResNet.
We investigated scaling parameters to optimize a model for ECG classifications.
We first present the fundamental architecture of ResNet in detail\footnote{We only consider one-dimensional layers, such as one-dimensional convolution, one-dimensional batch normalization, and so on, because ECG is a type of time-series data.}.

\subsection{Fundamental Architecture}
Equation~\ref{eq1} shows the architecture of ResNet.

\begin{equation}
\label{eq1}
y = (\text{FC} \circ \text{GAP} \circ R_4 \circ R_3 \circ R_2 \circ R_1 \circ S)(x),
\end{equation}
where $S$ is a single stem block and $(R_n)$ is a residual block, \textit{GAP} is a global average pooling layer, and \textit{FC} is a fully connected layer. 

The stem block $S$ is composed of a single unit function $F$ with a max pooling layer, as illustrated in Equation~\ref{eq2}. The function $F$ consists of a one-dimensional convolution, batch normalization, and ReLU activation function. Given that there are 12 leads in an ECG, the number of input channels for convolution is set at 12 (i.e., $w \in \mathbb{R}^{kernel \times 12 \times C_{out}}$).

\begin{gather}
\label{eq2}
F(x,w) = \sigma(\text{BN}(\text{Conv}(x,w)))\nonumber , \quad w \in \mathbb{R}^{kernel \times C_{in} \times C_{out}}\\
S = \text{Pool}(F(x,w)))),
\end{gather}
where $x$ represents the input and $w$ denotes convolution weights, which are defined by the product of the kernel size ($kernel$), the number of input channels ($C_{in}$), and the number of output channels ($C_{out}$). \textit{Conv} consists of a single one-dimensional convolution, \textit{BN} refers to a batch normalization layer, $\sigma$ is a ReLU activation function, and \textit{Pool} is a max pooling layer.

The residual block $R$ consists of multiple residual layer $L$ as shown in Equation~\ref{eq3} 
\begin{equation}
\label{eq3}
\centering
R = (L_d \circ \cdot \cdot \cdot \circ L_1) (x),
\end{equation}
where $d$ is the layer depth. 

Equation~\ref{eq4} illustrates the structure of $L$, which is composed of two sequential function $F$. In the first residual layer $L$, except for the initial residual block, the first function $F$ doubles the number of output channels through a convolution \textit{Conv} compared to the number of input channels. Subsequent $F$s maintain an equal number of input and output channels. The input $x$ is skip-connected with the output after the second batch normalization layer and then passed through the activation function. Prior to the skip-connection, the initial $L$ in $R_n$ processes $x$ using \textit{Pool} and \textit{Conv}, while the remaining layers perform identity mapping.

\begin{equation}
\label{eq4}
L = \\
\begin{cases}
    F(F(x,w)),2w)+\text{Pool}(\text{Conv}(x))& \text{if } n>1, d=1\\ 
    F(F(x,2w)),2w) + x& \text{otherwise}\\
\end{cases}
\end{equation}
where $w$ is convolution weights, $n$ is an index of blocks, and $d$ is an index of layers in a block, respectively.

\subsection{Scaling Parameters on Neural Network}

There are three hyperparameters that influence the network scale: the depth of layers (\textit{depth} $D$), the number of convolution channels (\textit{channels} $C$), and the size of the convolution kernel (\textit{kernel size} $K$). To examine the impact of these scaling parameters on optimization performance, we first established their search spaces. For the number of layers (layer depth $D$), we considered the minimum $D$ as two, based on ResNet18 (\cite{he2016deep}), the smallest variation among the ResNet models. We thus set the search space for the depth of residual layers to $D\in {2,4,8,16}$, where each corresponds to a total of 18, 34, 66, and 130 convolution layers, respectively.

For ECG analysis as well as tasks in the computer vision domain, small odd numbers are typically used as the kernel size $K$. The smallest kernel size is three, so we set the range to $K \in {3,5,9,15}$.

For the number of channels ($C$), most neural networks typically set the final number of output channels to 512 (\cite{he2016deep, elias2022deep}). We were curious about how adjusting the number of channels would affect the performance on ECG classifications. We considered the search space for the final output channels to be 128, 256, 512, and 1024, which corresponds to $C \in {16, 32, 64, 128}$.

\begin{table}[h]
\centering
\caption{Network scale of the fundamental architecture. $K$ denotes kernel size, $C$ does the number of channels, $d$ does the depth of residual layers, and $length$ does a input length of ECGs}
\begin{tabular}{c|c|c}
\toprule
  & Output size & Weights\\
 \midrule
 Input &  $length \times 12$ & \\ 
 $S$ & $length/4 \times C$  &
 \begin{tabular}[c]{@{}c@{}} $K \times 12 \times C$ \end{tabular}  \\
 $R_1$ & $ length/8 \times C$ & 
  \Big[ \begin{tabular}[c]{@{}c@{}} $ \begin{cases}
K \times C \times C & \text{if } d=1\\ 
K \times C \times C & \text{otherwise}
\end{cases} + K \times C \times C
$ \end{tabular} \Big] $\times D $ \\
 $R_2$ & $length/16 \times 2C$ & 
  \Big[ \begin{tabular}[c]{@{}c@{}} $ \begin{cases}
K \times C \times 2C & \text{if } d=1\\ 
K \times 2C \times 2C & \text{otherwise}
\end{cases} + K \times 2C \times 2C
$ \end{tabular} \Big] $\times D$ \\
 $R_3$ & $length/32 \times 4C$ & 
  \Big[ \begin{tabular}[c]{@{}c@{}} $ \begin{cases}
K \times 2C \times 4C & \text{if } d=1\\ 
K \times 4C \times 4C & \text{otherwise}
\end{cases} + K \times 4C \times 4C
$\end{tabular} \Big] $\times D$ \\
 $R_4$ & $length/64 \times 8C$ & 
  \Big[ \begin{tabular}[c]{@{}c@{}} $ \begin{cases}
K \times 4C \times 8C & \text{if } d=1\\ 
K \times 8C \times 8C & \text{otherwise}
\end{cases} + K \times 8C \times 8C
$\end{tabular} \Big] $\times D$  \\ 
 GAP & $ 8C$ & \\
 FC & \# of labels & $ 8C \times \text{\# of labels}$ \\
\bottomrule
\end{tabular}
\label{tab:framenet}
\end{table}

Table~\ref{tab:framenet} summarizes the network scale by layer depth $D$, kernel size $K$, and the number of channels $C$. $length$ denotes the input ECG length. The number of weights increases as the network becomes deeper. The first output has a quarter of the original length due to the striding of the convolution and a pooling layer at the stem block. As the input passes through the residual blocks, the length is halved while the number of channels is doubled. The number of labels depends on the labels of the ECG datasets.

\section{Experiment}

\subsection{Dataset}
We employed two ECG datasets in our experiment. The first dataset is Physionet Challenge 2021 (Physionet) dataset (\cite{reyna2021will}). This dataset contains standard 12-lead ECGs with cardiac arrhythmia labels. It is composed of seven databases collected from various institutions with unique demographics. In our experiments, we omitted two of the seven databases that had fewer than several hundred samples and focused on five databases: PTB-XL (\cite{wagner2020ptb}), CPSC (\cite{liu2018open}), Shaoxing (\cite{zheng202012}), G12EC (\cite{reyna2021will}), and Ningbo (\cite{zheng2020optimal}). Each database is collected from the United States (G12EC), Germany (PTB-XL), and China (CPSC, Shaoxing, Ningbo), with the signal length varying between 10 and 60 seconds. The total number of ECGs is approximately 88,000, and each ECG is associated with one or more labels among the 26 diagnostic classes, which cover a wide range of categories.
The second one is Alibaba Tianchi competition (Alibaba) dataset (\cite{tianchi}). The ECG signals have a duration of 10 seconds each, and the overall collection consists of approximately 20,000. Each ECG is tagged with one or more labels from 33 classes. Since a few class include extremely small number of class (e.g. only three ECGs are labeled as QRS Low Voltage), we limited our study to labels with more than prevalence of 0.1\%, and thus used 17 classes for experiments. 
More details of datasets are provided in Appendix~\ref{append:a}.

\subsection{Evaluation}
We use the macro-average F1 score as the evaluation metrics. The F1 score is a harmonic mean of precision and recall. In multi-label classification, each sample belongs to multiple classes. The macro-average F1 score calculates the F1 score for each class independently and then averages them. This approach treats all classes equally, regardless of class distribution, making it particularly useful for the imbalanced dataset.
We divided the dataset into train, validation, and test sets at a ratio of 0.7, 0.15, and 0.15, respectively.

\subsection{Training Procedure}
\label{sec3:procedure}
As ECGs are gathered with varying lengths and sampling rates, we standardize them. For signal preprocessing, we trimmed the ECG length to 10 seconds. If the length of the ECG is longer than 10 seconds, we randomly cropped it to a 10-second duration; otherwise, we extended the ECG to 10 seconds by padding with zeros. Additionally, we resampled all ECGs to a sampling rate of 250Hz.
In addition, for the Alibaba dataset, out of the standard 12 leads of an ECG, this dataset does not contain leads III, aVR, aVL, and aVF. Thus, by applying the principles of Einthoven's Triangle and Goldberger's augmented lead (\cite{mirvis2001electrocardiography}), we filled in the four missing leads using leads I and II.

Hyperparameter tuning is essential for selecting an appropriate model for a given task. Without hyperparameter tuning, the performance and generalizability of a model may be suboptimal, leading to unreliable conclusions~(\cite{bergstra2012random}). However, computational limitations have prevented previous studies from exploring a wide range of hyperparameters (\cite{strodthoff2020deep}). For example, one of the previous studies on ECG classification benchmark only searched 9 sets of learning rate and batch size (\cite{nonaka2021depth}), and the majority of studies do not present the precise procedure of the hyperparameter optimization.

In this study, we explored 50 combinations of hyperparameters for each set of scaling parameters, thereby resolving a limitation of previous studies.
Tuned hyperparameters consist of learning rate, weight decay, dropout, and other regularization techniques.
The search space used in the experiment is presented in Table~\ref{tab:space}.
In training, we employed the Adam optimizer (\cite{kingma2014adam}) and a one-cycle learning rate scheduler (\cite{smith2019super}) that peaks at epoch 10.
The batch size was fixed at 512.
We randomly sampled the learning rate $\in [10^{-4}, 10^{-2}]$ and weight decay $\in [10^{-6}, 10^{-4}]$ for the optimizer. The dropout rate was randomly chosen from $0$ to $0.3$ in increments of $0.05$.
We applied 18 ECG data augmentation methods (\cite{lee2022efficient}) using the RandAugment policy (\cite{cubuk2020randaugment}). The number of augmentation methods was randomly selected from ${0, 1, 2}$, and their intensity was chosen from 10 levels. Furthermore, we employed Mixup [\cite{zhang2017mixup}], randomly selecting a beta distribution from $\beta \in \{0.0, 0.1, 0.2\}$.
During hyperparameter searching, we used the asynchronous successive halving algorithm (ASHA) with a grace period of 10 and a reduction factor of 2 (\cite{li2018massively}).
Our training procedure is implemented with Ray (\cite{moritz2018ray}).

\begin{table}[ht]
\centering
\caption{Hyperparameter Search Space}
\vspace{3mm}
\label{tab:hyperparams}
\begin{tabular}{l | l}
\toprule
Hyperparameter & Search Space \\ 
\midrule
Learning rate & [$10^{-4}, 10^{-2}$] \\
Weight decay & [$10^{-6}, 10^{-4}$] \\
Dropout rate & \{0, 0.05, 0.1, 0.15, 0.2, 0.25, 0.3\} \\
Number of augmentation & \{0, 1, 2\} \\
Magnitude of augmentation & \{0, 1, 2, 3, 4, 5, 6, 7, 8, 9, 10\} \\
Beta distribution of Mixup & \{0, 0.1, 0.2\} \\
\bottomrule
\end{tabular}
\label{tab:space}
\end{table}

\section{Result}
\label{sec:result}
\subsection{Impact of Optimization for Scaling Parameter on ECG classification} 

\begin{figure*}[h]
    \centering

    \includegraphics[width=1.0\textwidth]{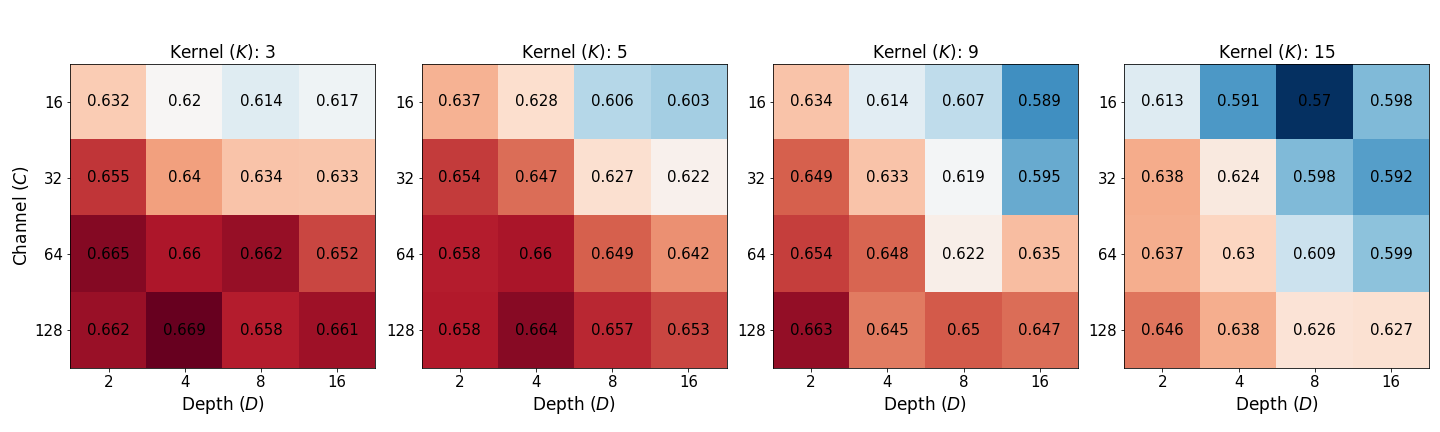}
    \includegraphics[width=1.0\textwidth]{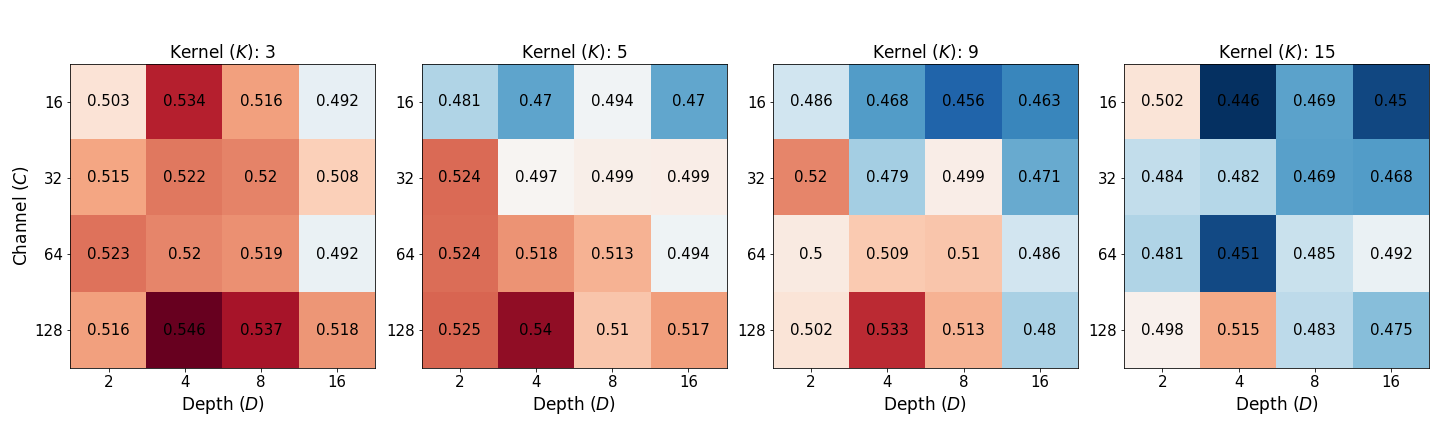}
    \caption{
    The performance of residual networks with different layer depth ($D$), number of channels ($C$), and kernel size ($K$) for the ECG classification. The first row is the result of Physionet 2021 challenge dataset~(\cite{reyna2021will}) and the second row is the result from Alibaba Tianchi competition dataset~(\cite{tianchi}). The scores in boxes are the average of F-1 scores on the multi-label classification. The more red, the higher performance, while the more blue, the lower performance.
    }
    \label{fig:performance}
\end{figure*}

Figure~\ref{fig:performance} shows the F1 score of the ECG classification for ResNet with different layer depths ($D$), numbers of channels ($C$), and kernel sizes ($K$). The selected hyperparameters for each network are provided in Appendix~\ref{append:b}. In the box color, the more red it is, the higher the performance, while the more blue it is, the lower the performance.

\textbf{Layer depth $\bf{(D)}$}: We observed a general trend in which the performance tends to improve as the depth $D$ decreases from both Physionet 2021 and Alibaba datasets. The trend is found regardless of kernel size $K$ and channel $C$. This findings clearly exhibit a distinct different pattern compared to computer vision domain, where it is known that performance increases as networks become deeper.
On the other hand, it is also important to note that a shallower layer depth is not always advantageous. Depending on the combinations of kernel size $K$ and the number of channels $C$, the optimal setting of layer depth can be different. 


\textbf{Number of channels $\bf{(C)}$}: We found a positive correlation between the number of channels $C$ and performance in both Physionet 2021 and Alibaba datasets. In the majority of cases, as we varied kernel size $K$ and layer depth $D$, the performance improved when the number of channels $C$ increased. In fact, for both Physionet and Alibaba datasets, all the best performances were observed in models with the largest number of channels irrespective to combinations of kernel size $K$ and layer depth $D$. This result is consistent with the established knowledge in the computer vision domain, indicating that wider networks are particularly well-suited for ECG classification tasks. 

\textbf{Kernel size $\bf{(K)}$}: The panels on the left side of Figure~\ref{fig:performance} show a more reddish hue than those on the right side in both Physionet 2021 and Alibaba datasets. This indicates that performance tends to decrease as the kernel size $K$ increases. Although recent studies in the computer vision domain have reported a positive impact of larger kernels on model performance [\cite{ding2022scaling}], our experiments for ECG classification show a contrasting trend, with larger kernels leading to reduced performance. 

In Physionet 2021 dataset, the best combination for ECG classification performance is a layer depth $D$ of 4, a number of channels $C$ of 128, and a kernel size $K$ of 3. In contrast, a combination with a layer depth $D$ of 8, a number of channels $C$ of 16, and a kernel size $K$ of 15 results in the worst performance. 
Although the size of dataset and label distribution in Alibaba differ from those of Physionet, the best and worst combination in Alibaba dataset were remarkably similar to that of Physionet 2021 one. The optimal configuration for Alibaba dataset and Physionet 2021 are identical. The worst performance model is generated when a layer depth $D$ of 4, a number of channels $C$ of 16, and a kernel size $K$ of 15.
Overall, our findings provide the following scaling parameter setting guide to improve performance in ECG classification tasks, regardless of datasets: a shallower network, a larger number of channels, and a smaller kernel size.

\subsection{Infusing our insight into hyperparameter optimization}
In practice, hyperparameter optimization is essential for achieving satisfactory performance in a given task (\cite{Jaderberg2017PopulationBT}).
The efficiency of hyperparameter optimization is heavily influenced by the predefined search space. A vast search space might not guarantee the best performance and could lead to increased time and resource consumption. Thus, it is essential to establish a well-designed range for the search space, which ensures efficient hyperparameter optimization and ultimately results in improved model performance.

Incorporating insights from our experiments can help researchers increase their chances of achieving optimal performance. To demonstrate this, we conducted additional experiments, modifying the range of scaling hyperparameters, specifically layer depth $D$, the number of channels $C$, and kernel size $K$. Table~\ref{tab:hpo-perf} shows the best performance depending on the size of search spaces, defining three distinct ranges: Large, Medium, and Optimal. The \textit{Large} range represents a wide hyperparameter search space, while the \textit{Medium} range is half of the \textit{Large} range, reflecting our findings of a shallower network, smaller kernel size, and larger number of channels. The \textit{Optimal} range includes settings that yield the best performance, as shown in Section~\ref{sec:result}. All other hyperparameters are configured as mentioned in Section~\ref{sec3:procedure}.

For the experiments, we first sampled 50 combinations in a search space. Then, we selected the best model by training and evaluating them. As a result, for both Physionet 2021 and Alibaba datasets, the performance improvement are observed by reducing the space to $1/8$ (\textit{Medium}) and $1/64$ (\textit{Optimal}) compared with \textit{Large}, respectively. Larger search spaces make it more challenging to find the optimal hyperparameters within the given combinations, potentially leading to suboptimal model performance.
These results demonstrate that models acquired through more precise hyperparameters outperform those obtained using a wider search space.

\begin{table}[h]
\centering  
\caption{
    Performance depending on the search space of hyperparameter optimization. \textit{\# Avail. comb.} denotes the number of available combinations. \textit{F1} represents the best F1 score in the sampled 50 combinations. The selected hyperparameters for each experiment are provided in Appendix~\ref{append:c}.
}
    \vspace{3mm}
    \begin{tabular}{@{}c|ccc|c|cc@{}}
    \toprule
    Physionet  & Depth ($D$)    & Channel ($C$)   & Kernel ($K$)  & \# Avail. comb. & F1  \\ \midrule
    Large  & \{2,4,6,8\} & \{16,32,64,128\} & \{3,5,9,16\} & 64 & 0.612      \\ 
    Medium & \{2,4\}        & \{64,128\}         & \{3,5\}  &  8   & 0.644 &    \\ 
    Optimal  & \{4\}           & \{128\}             & \{3\} &   1    & 0.669 &    \\ \bottomrule
    \multicolumn{7}{c}{}\\
    \toprule
    Alibaba & Depth ($D$)    & Channel ($C$)   & Kernel ($K$)  & \# Avail. comb. & F1  \\ \midrule
    Large  & \{2,4,6,8\} & \{16,32,64,128\} & \{3,5,9,16\} & 64 & 0.000      \\ 
    Medium & \{2,4\}        & \{64,128\}         & \{3,5\}  &  8   & 0.000 &    \\ 
    Optimal  & \{4\}           & \{128\}             & \{3\} &   1    & 0.000 &    \\ \bottomrule
    \end{tabular}    

\label{tab:hpo-perf}
\label{tab:results}
\end{table}



\section{Discussion}

\subsection{Sub-group Analysis}
As described in Table \ref{tab:label-physionet} and Table \ref{tab:label-Alibaba}, both Physionet 2021 and Alibaba datasets present highly imbalances in their distribution. 
To determine if our experimental findings remain consistent across the study population, we conducted a sub-group analysis.
The Physionet 2021 dataset is composted of various classes and databases, where each of them has its own distict distribution.
For example, the ratio between the NSR and Brady classes is approximately 100:1, and 65\% of NSR labels are from the Ningbo database, while 67\% of SB labels are from the G12EC database. Additionally, the distribution of labels across the database sources is uneven, with no database containing all 26 classes, and some databases contain fewer classes than others.
Similartly, the Alibaba dataset also exhibits substantial imbalances. Althought it is collected from a single source, the skewness of class distribution is comparable to the multi-sourced Physionet 2021 dataset.
For example, the prevalence of NSR approaches nealy 50\%, while the prevalence of NSTAb is only 0.03\%.

\subsubsection{Performance Depending on Labels}

\begin{figure}
    \vspace{-15mm}
    \centering
    \includegraphics[width=0.8\textwidth]{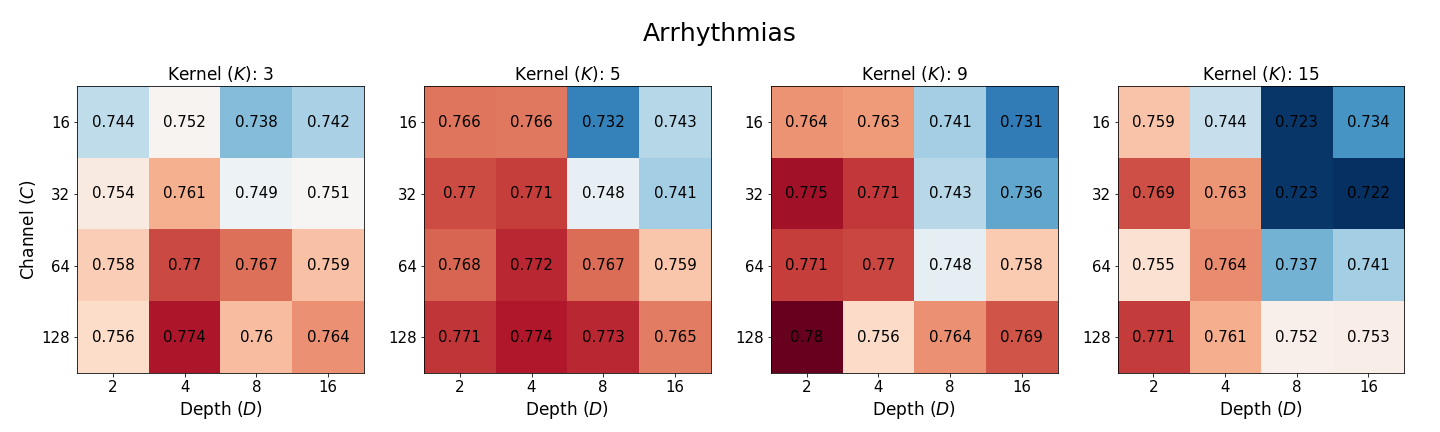}
    \vspace{-1mm}
    \includegraphics[width=0.8\textwidth]{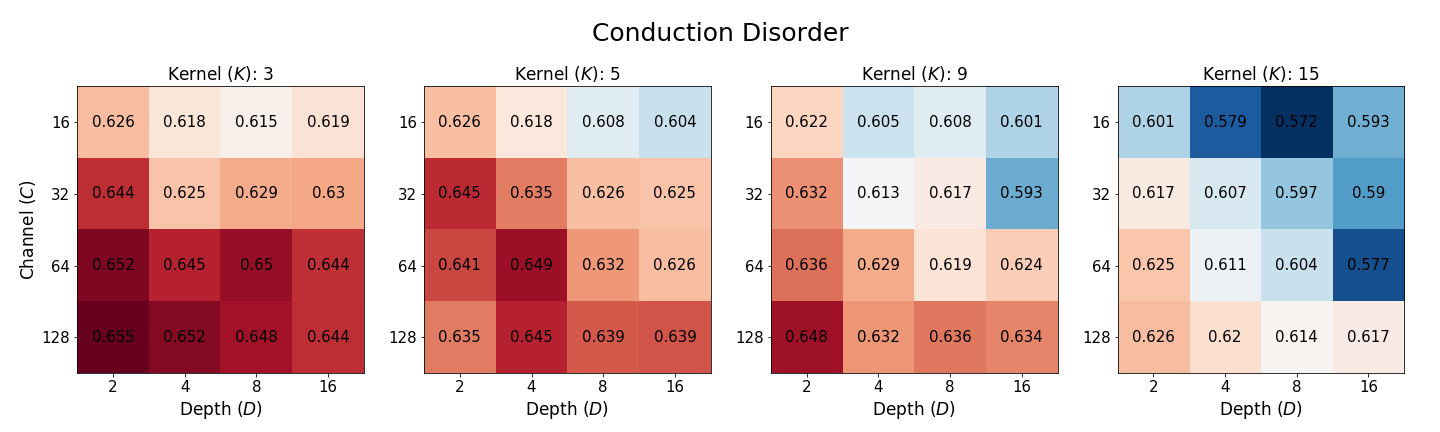}
    \vspace{-1mm}
    \includegraphics[width=0.8\textwidth]{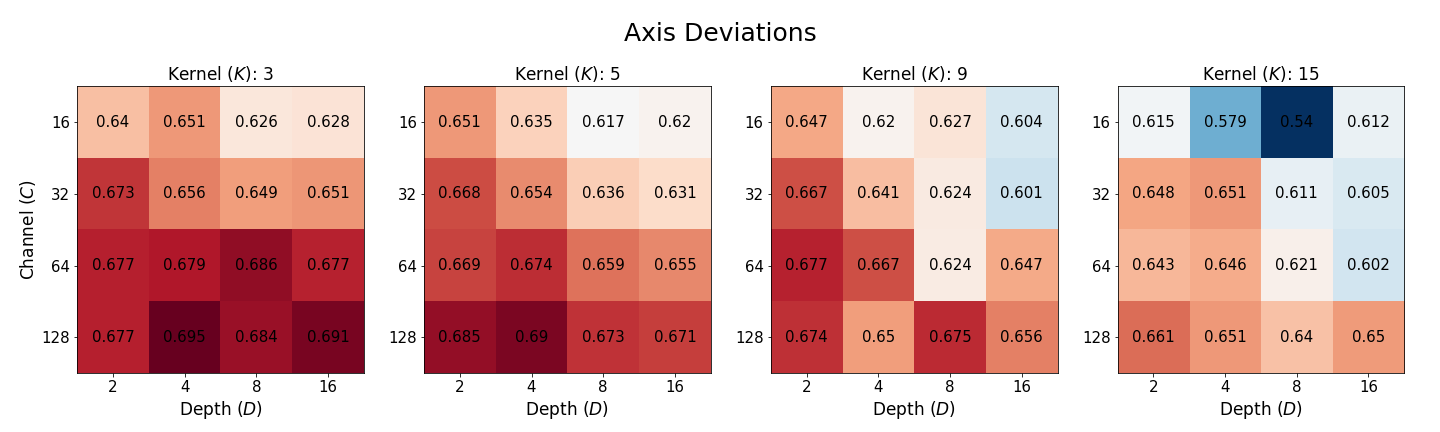}
    \vspace{-1mm}
    \includegraphics[width=0.8\textwidth]{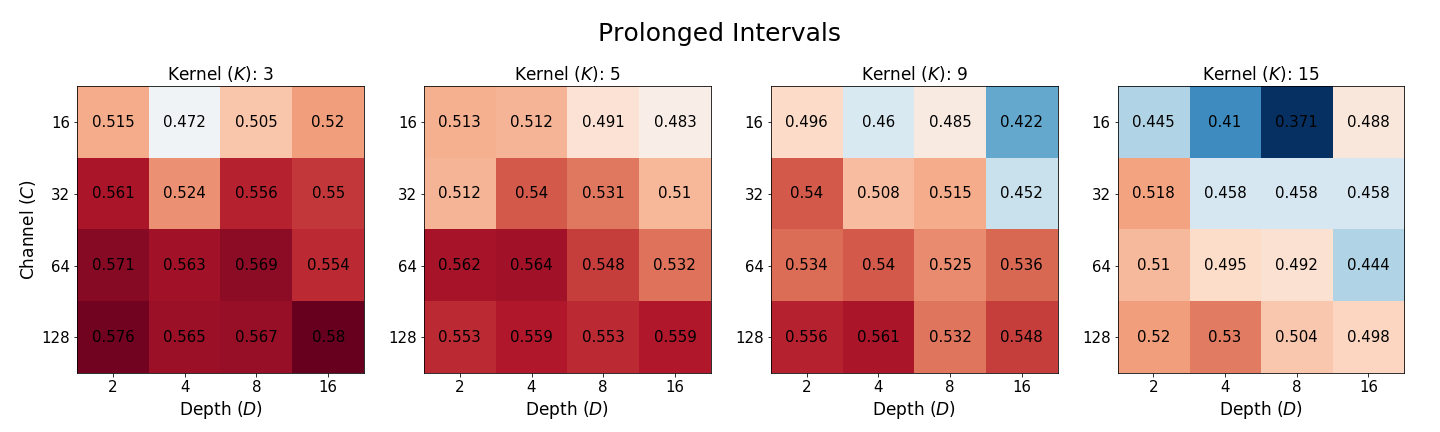}
    \vspace{-1mm}
    \includegraphics[width=0.8\textwidth]{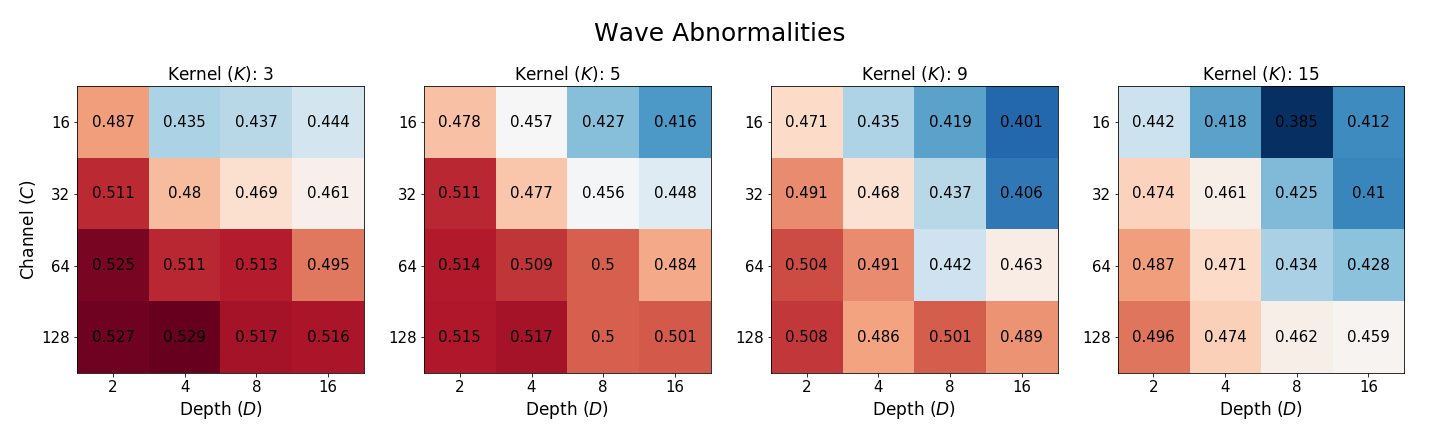}
    \vspace{-1mm}

    \includegraphics[width=0.8\textwidth]{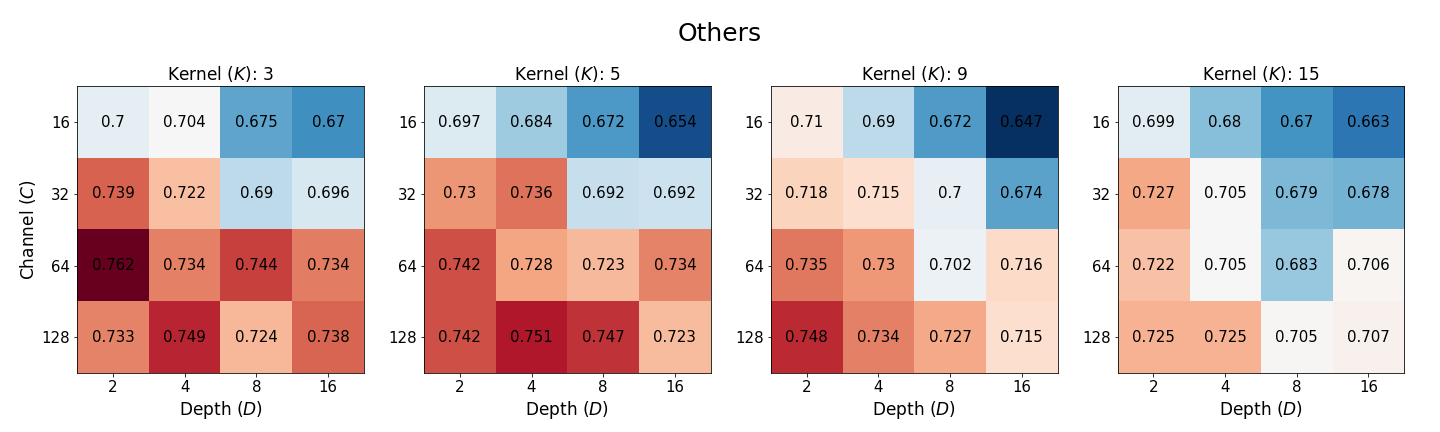}
    \caption{Classification performance of networks with different layer depth ($D$), the  number of channels ($C$), and kernel size ($K$) on six label categories in Physionet2021 dataset.}

\label{fig:performance-physionet-category}
\end{figure}

\begin{figure}
    \centering
    \includegraphics[width=0.8\textwidth]{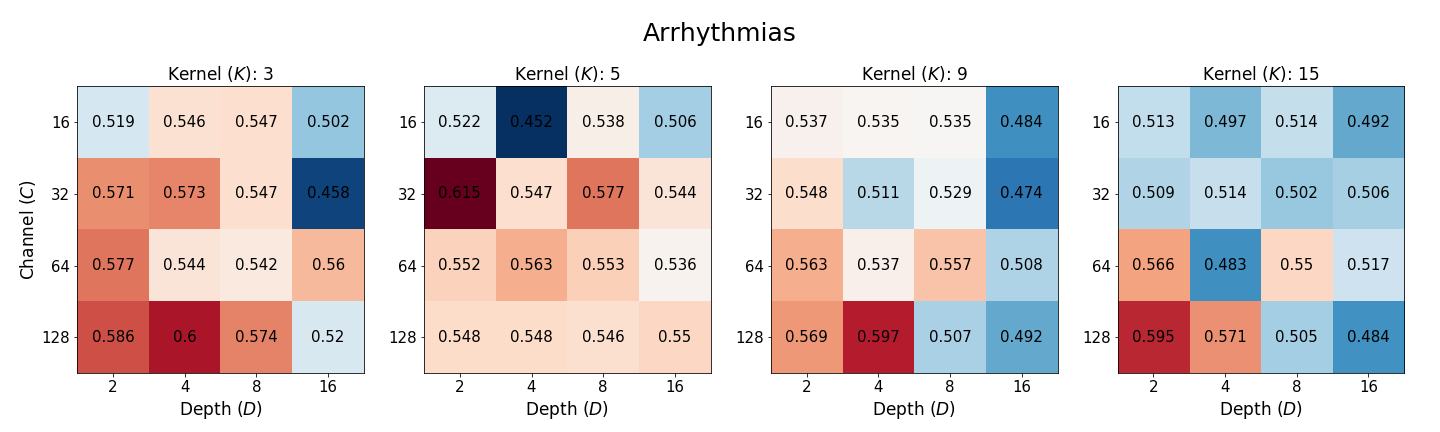}
    \vspace{-1mm}
    \includegraphics[width=0.8\textwidth]{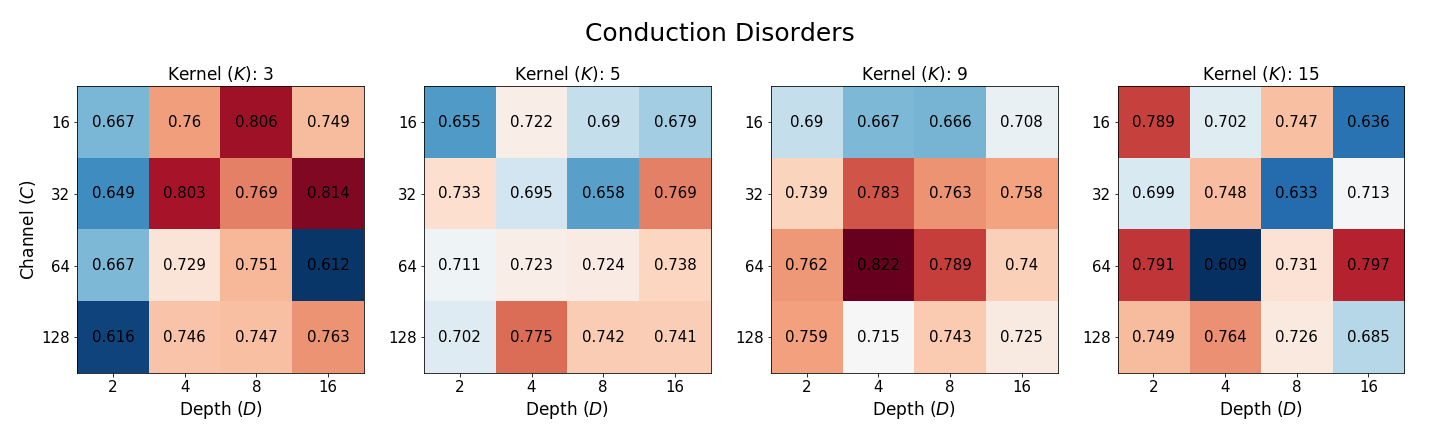}
    \vspace{-1mm}
    \includegraphics[width=0.8\textwidth]{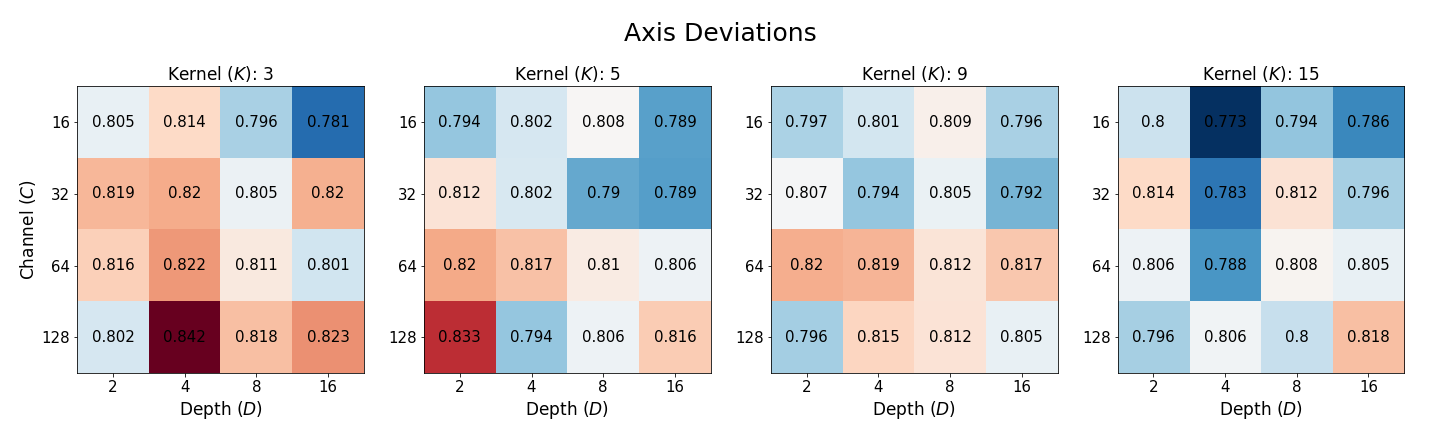}
    \vspace{-1mm}
    \includegraphics[width=0.8\textwidth]{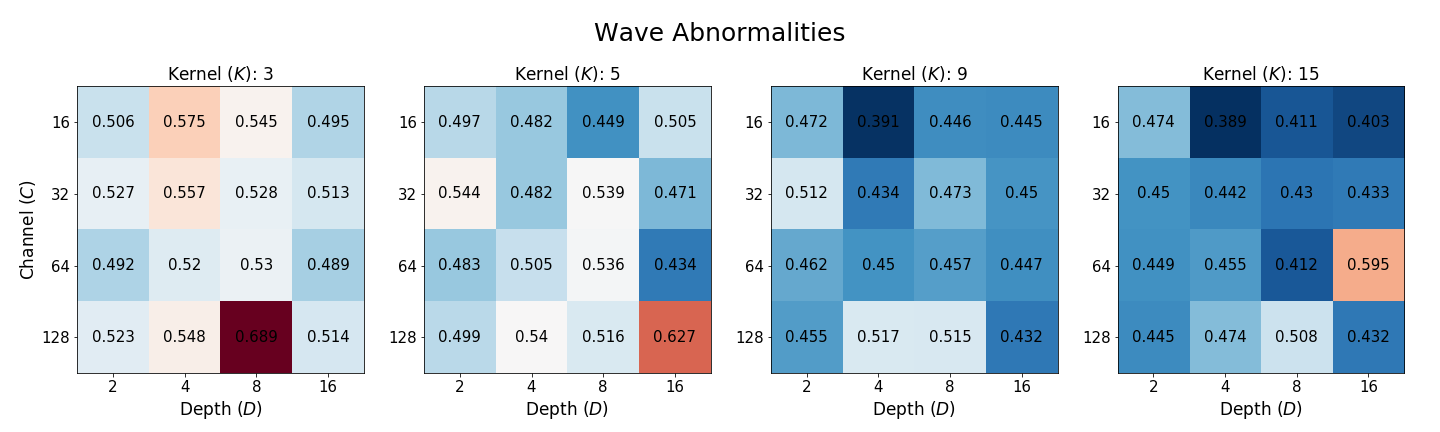}
    \vspace{-1mm}

    \includegraphics[width=0.8\textwidth]{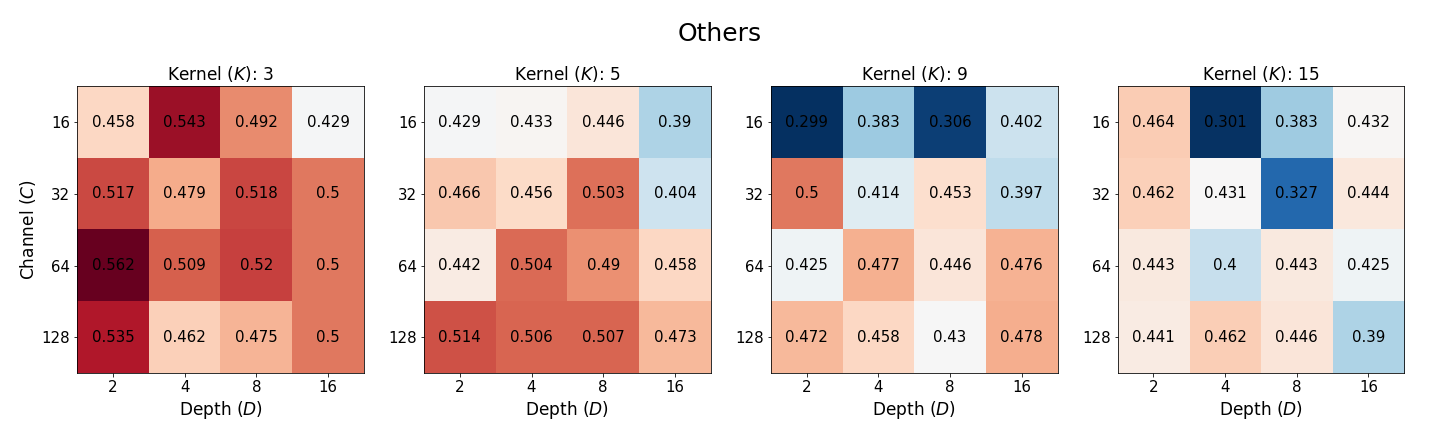}
    \caption{Classification performance of networks with different layer depth ($D$), the  number of channels ($C$), and kernel size ($K$) on six label categories in Alibaba dataset.}

\label{fig:performance-Alibaba-category}
\end{figure}

Figure~\ref{fig:performance-physionet-category} and Figure~\ref{fig:performance-Alibaba-category} demonstrate the performance of the model across label categories in Physionet 2021 and Alibaba datasets. 
In the case of Physionet 2021 dataset, the performance changes in each category are similar to the performance changes in entire classes. The optimal combination ($D4$-$C128$-$K3$) derived from the experiment in Section~\ref{sec:result} may not belong to the best performance, but it remains among the top-performing models across all categories. This consistency suggests that the proposed scaling method is effective across all categories. 

However, in the Alibaba dataset, while the overall trend of performance regarding to depth, width, and kernel size remains consistent across all categories, the degree of similarity is substantially lower than in the Physionet 2021 dataset. Specifically, there are numerous outliers that contradict the general trend, such as $C$4, $D$2, $K$5 and $C$64, $D$4, $K$9.
One possible explanation for this could be attributed to the number of labels. Unlike the Physionet 2021 dataset, the prevalence of labels within the Alibaba dataset is relatively low; only three out of the seventeen classes exhibit a prevalence of 10\% or higher. This significant class imbalance could potentially lead to unstable results. 
Appendix~\ref{append:a} provides a more comprehensive analysis of the model performance on specific labels. All performances depending on each label among all available combinations are described in detail in Appendix~\ref{append:d}.

\subsubsection{Performance Depending on Database Sources}

\begin{figure}
    \centering
        \includegraphics[width=0.8\textwidth]{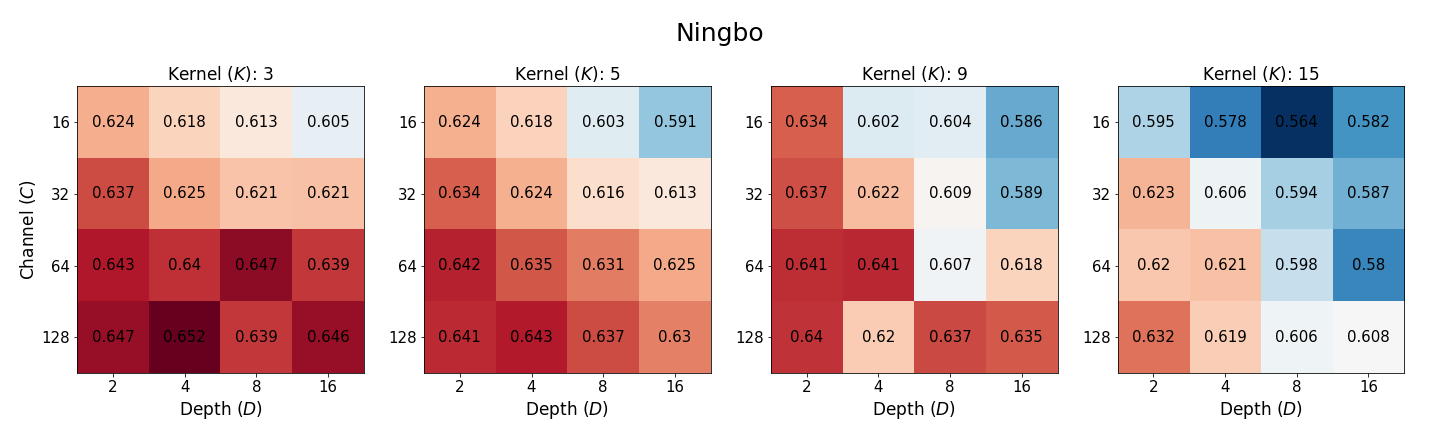}
        \vspace{-1mm}
        \includegraphics[width=0.8\textwidth]{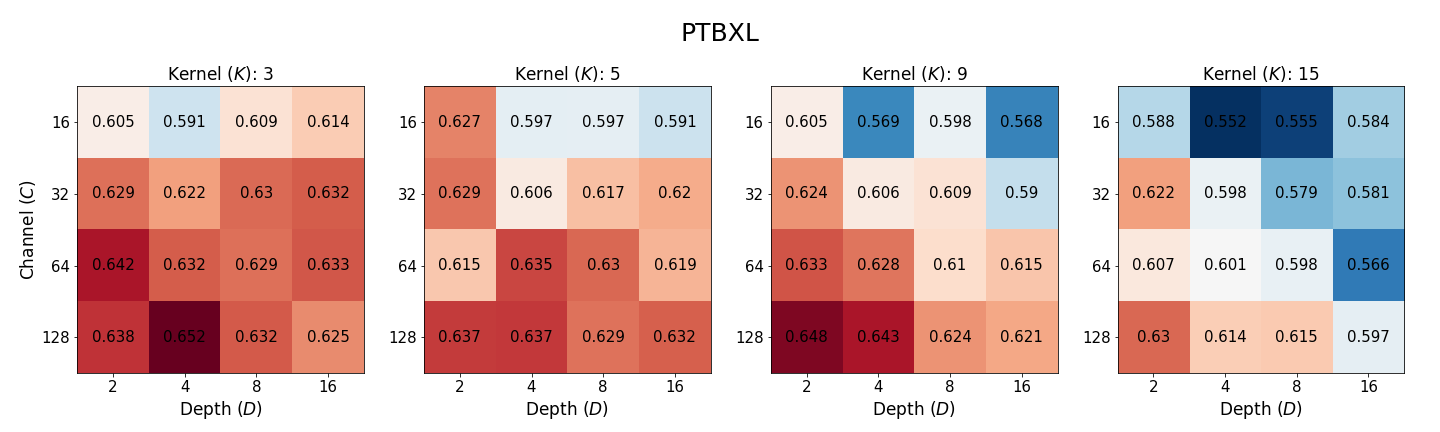}
        \vspace{-1mm}
        \includegraphics[width=0.8\textwidth]{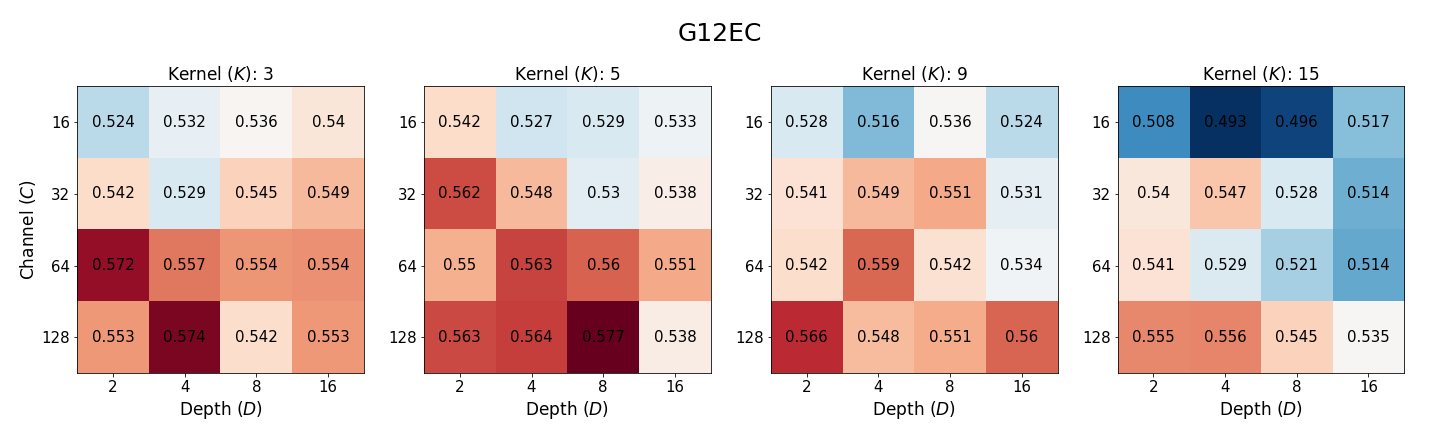}
        \vspace{-1mm}
        \includegraphics[width=0.8\textwidth]{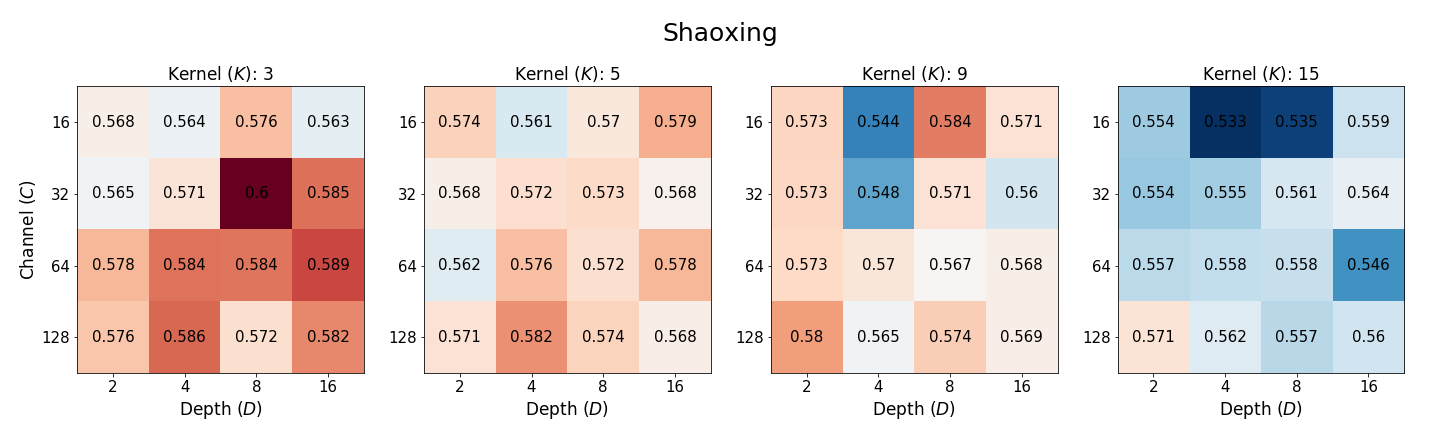}
        \vspace{-1mm}
        \includegraphics[width=0.8\textwidth]{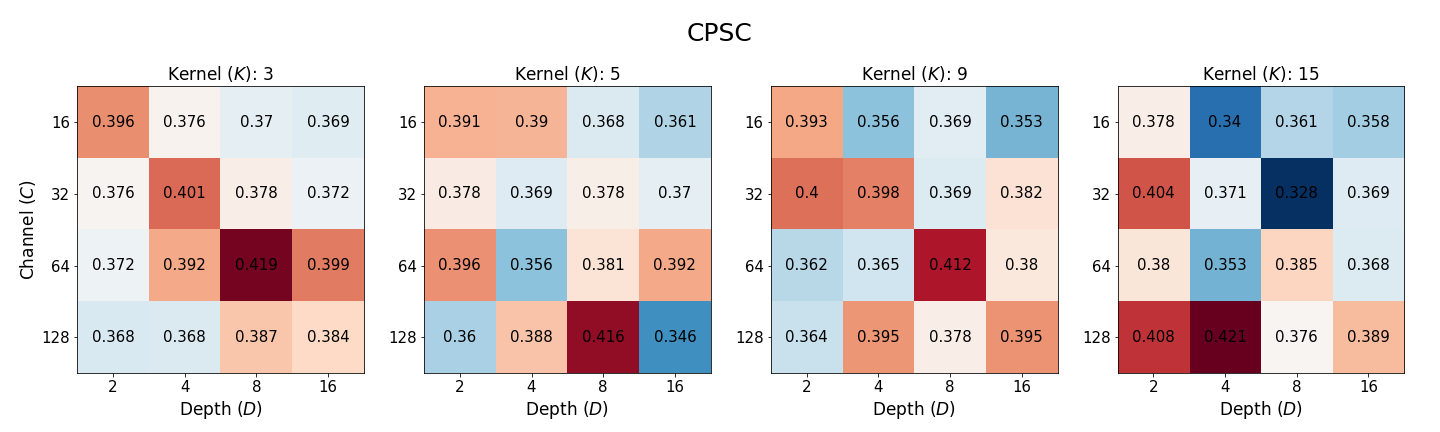}
\caption{Classification performance of networks with different layer depth ($D$), the number of channels ($C$), and kernel size ($K$) depending on database sources.}
\label{fig:performance-population}
\end{figure}

Figure~\ref{fig:performance-population} shows the model performance according to the database sources in Physionet 2021 dataset. It appears that results in the Ningbo, PTBXL, G12EC, and Shaoxing databases are consistent with the entire dataset. On the other hand, we observe the different performance trend in CPSC database. 
One possible reason is that the CPSC database has some rare labels such as NSIVCB, PR, SA, Brady, TAb, and TInv, where a small number of samples for these classes can significantly impact the F1 scores since we used the macro-averaged F1 score as a metric.


\subsection{Correlation between Scaling Parameters and Performance}
Previous research in the computer vision domain has shown that there is a positive correlation between model performance and all scaling parameters  (\cite{peng2017large, tan2019efficientnet, ding2022scaling}). Specifically, deeper residual layers, larger kernel sizes, and many channels can contribute to better performance.

However, our findings from ECG analysis provide a different result on the correlation between model performance and scaling parameters compared to the computer vision domain. Table~\ref{tab:corr} shows the correlation between neural networks with different scales and their performance in both ECG analysis and computer vision. 
Contrary to the expectations, we observed a negative correlation between performance and both layer depth $D$ and kernel size $K$. Only the number of channels $C$ showed a positive correlation with performance, consistent with the observatoins in the computer vision.

We conjecture that observed discrepancy is attributed to the periodic nature of ECG signals and their effect on the receptive field and global average pooling(GAP).
As illustrated in Figure~\ref{fig:ecg}, the patterns in each ECG cycle show consistent and repeat over the time. 
Because of this periodicity, a receptive field that spans only a few second of ECG (i.e. one or two cycle of ECG) is sufficient for extracting the essential information from the signal. 
Thus, in contrast to computer vision, where the large receptive field is preferred for capturing global feature, the advantage of large receptive field is relatively insignificant in ECG classification.
Further, the GAP following small receptive field functions as an ensemble of different view of signal, providing generalized and robust representation.
On the other hand, a large receptive field that covers the entire signal does not benefit as much from GAP as a small receptive field. (See Appendix~\ref{append:e} for further discussion). 

To investigate whether the influence of periodicity persists when it is removed, we cropped ECG signals into 2-second and 1-second segments to eliminate the effect of varying heart rates in Physionet 2021 dataset. This resulted in approximately two cycles present in a 2-second signal and one cycle in a 1-second signal, respectively. The correlation between network scaling and performance for the cropped signals is shown in the 2s and 1s at ECG columns in Table~\ref{tab:corr}, respectively. Specifically, we found that the depth of layers led to a near-complete loss of correlation, while the kernel size was less affected. In contrast, the correlation remained high for the number of channels. While our experimental results cannot fully explain our conjecture, they support the idea that the periodicity of ECG signals has some influence on network scaling decisions.

\begin{table}[h]

\centering  
\caption{
    Correlation between the neural network's scaling parameters and its performance based on ECG length.
}
\begin{tabular}{c|c|rrr}
\toprule
\multicolumn{1}{l|}{\multirow{2}{*}{}} & \multirow{2}{*}{\begin{tabular}[c]{@{}c@{}}Computer\\ Vision\end{tabular}} & \multicolumn{3}{c}{ECG}                                                   \\
\multicolumn{1}{l|}{}                  &                         & \multicolumn{1}{c}{10s} & \multicolumn{1}{c}{2s} & \multicolumn{1}{c}{1s} \\ \hline
Layer depth (D)                              & Positive (+)            & -0.83                   & -0.62                  & -0.2                   \\
Kernel size (K)                             & Positive (+)            & -0.92                   & -0.84                  & -0.68                  \\
Number of channels (C)                            & Positive (+)            & 0.93                    & 0.92                   & 0.87               \\ \bottomrule   
\end{tabular}
\label{tab:corr}
\end{table}

\section{Conclusion}
In recent studies, CNN-based models, particularly ResNet, have been widely used to achieve higher performance in ECG classifications. However, the optimization of network scaling, which is a vital process for creating efficient and precise models, has often been overlooked. To address this, we examined the effects of different combinations of primary scaling parameters, such as network depth, convolution kernel size, and the number of channels, on the Physionet2021 dataset and Alibaba dataset.

Our results show that performance enhancement can be achieved by employing models with shallower networks, a greater number of channels, and smaller convolution kernel sizes. This finding is crucial because optimizing network scaling can be both time-consuming and computationally demanding. By refining the search space for scaling parameters based on our results, researchers can more effectively reach optimal outcomes within a given time.

Moreover, we investigated the performance variations by network scaling in relation to the dataset's labels and inferred the rationale behind our findings. This research provides a better understanding of network scaling optimization in ECG analysis and empowers researchers to develop more efficient and effective models for ECG classification.

\pagebreak

\bibliography{refs}

\pagebreak

\appendix

\setcounter{table}{0}
\renewcommand{\thetable}{A.\arabic{table}}
\setcounter{figure}{0}
\renewcommand{\thefigure}{A.\arabic{figure}}

\section{Architecture of Deep learning models on ECG classification}
\label{append:review}

The following table shows the scaling parameters of the models that were developed in the previous studies. The papers in the table solely used ResNet for ECG classification, along with different parameters such as the depth of the layers, the number of channels, and the dimensions of the kernel. We noticed a lack of consensus on the ideal strategy for optimizing the scale of these networks. Even though the studies by \cite{vazquez2021two, han2021towards} and \cite{sakli2022resnet} used the same datasets, their network architectures were different.

\begin{table}[ht]
\caption{Comparison of Scality of ResNet for ECG Classification}
\centering
\small

\begin{tabular}{l|c|c|c|c}
\hline
 & \textbf{Year} & \textbf{Layer Depth} & \textbf{Number of channels} & \textbf{Kernel size} \\

\toprule

\cite{pourbabaee2017deep} & 2017 & 5 & 64 & 32 \\
\cite{hannun2019cardiologist} & 2019 & 34 & 512 & 16 \\
\cite{hsieh2020detection} & 2020 & 13 & 512 & 5 \\
\cite{ribeiro2020automatic} & 2020 & 10 & 320 & 16 \\
\cite{vazquez2021two} & 2021 & 7 & 256 & 5 \\
\cite{han2021towards} & 2021 & 14 & 128 & 15 \\
\cite{park2022study} & 2022 & 152 & 2048 & 3 \\
\cite{sakli2022resnet} & 2022 & 50 & 512 & 3 \\
\bottomrule
\end{tabular}
\end{table}

\section{Details of datasets}
\label{append:a}
Both Physionet2021 and Tinachi datasets contains more than 20 label classes. We categorize the labels into six category based on their condition: Arrhythmia, conduction disorder, axis deviation, prolonged intervals, wave abnormalities, and the others. Each category represents the following conditions:

\begin{itemize}
    \item Arrhythmia is an irregularity in the heart's rhythm due to problems in the heart's electrical conduction system. They can be classified into several types, from atrial fibrillation, atrial flutter, ventricular tachycardia, and ventricular fibrillation to sinus premature atrial contraction and premature ventricular contracction. 
    
    \item Conduction disorder occurs when the heart's electrical signals are delayed or blocked. Examples include atrioventricular (AV) block, and various bundle branch block, such as Left Bundle Branch Block, Right Bundle Branch Block, and 1st Degree AV Block.
    
    \item Axis deviation refers to an abnormal orientation of the heart's electrical axis, which can be assessed by analyzing the QRS complex on an electrocardiogram (ECG). Axis deviation can be categorized as left, right, or indeterminate. Also, we include ECG with lead displacement in this category.
    
    \item Prolonged intervals are abnormalities in the duration of specific ECG segments or intervals, such as the PR, QRS, or QT intervals. Both short and long intervals are included in this category.
    
    \item Wave involves changes in the size, shape, or polarity of ECG waves, such as the P wave, QRS complex, or T wave. Interval abnormalities with respect to magnitude is also included in this category, such as ST segment abnormalities.
    
    \item The other category contains condition that is not included any category described above. 
    This represent a normal or near-normal cardiac electrical activity or the presence of an artificial pacemaker generating the heart's rhythm. 
\end{itemize}

As shown in Table \ref{tab:label-physionet} and Table \ref{tab:label-Alibaba}, both datasets are highly skewed. Specifically, the average prevalence of labels in Physionet 2021 is 0., and the prevalence of labels in Alibaba is only 0.042. Further, in the case of Physionet 2021, the imbalance of distribution between sub-datasets is significant. For example, more than half of CPSC dataset has fewer than five labels for each class, whereas G12EC, Ningbo, and PTBXL contain relatively balanced number of classes.


\begin{table}[ht]
\centering
\caption{The number of labels in Physionet2021 dataset.} 
\small
\begin{tabular*}{\columnwidth}{@{\extracolsep{\fill}} l l r r r r r r}
\toprule
Category & Label & Ningbo & PTBXL & G12EC & Shaoxing & CPSC & Total \\
\midrule

Arrhythmia & AF & 1514 & 570 & 0 & 1780 & 1374 & 5238 \\
            & AFL & 73 & 186 & 7615 & 445 & 54 & 8373 \\
            & SA & 772 & 455 & 2550 & 0 & 11 & 3788 \\
            & SB & 637 & 1677 & 12670 & 3889 & 45 & 18918 \\
            & STach & 826 & 1261 & 5687 & 1568 & 303 & 9645 \\
            & PAC & 1063 & 555 & 640 & 258 & 740 & 3256 \\
            & PVC & 0 & 357 & 1091 & 294 & 194 & 1936 \\
\addlinespace
Conduction disorder 
            & BBB & 0 & 116 & 385 & 0 & 0 & 501 \\
            & LBBB & 536 & 231 & 248 & 205 & 274 & 1494 \\
            & RBBB & 542 & 556 & 1291 & 454 & 1971 & 4814 \\
            & 1AVB & 797 & 769 & 893 & 247 & 828 & 3534 \\
            & IRBBB & 1118 & 407 & 246 & 0 & 86 & 1857 \\
            & NSIVCB & 789 & 203 & 536 & 235 & 4 & 1767 \\
            & LAnFB & 1626 & 180 & 380 & 0 & 0 & 2186 \\
\addlinespace
Axis deviation 
            & LAD & 5146 & 940 & 1163 & 382 & 0 & 7631 \\
            & RAD & 343 & 83 & 638 & 215 & 1 & 1280 \\
\addlinespace
Prolonged interval
            & LPR & 340 & 0 & 40 & 12 & 0 & 392 \\
            & LQT & 118 & 1391 & 334 & 57 & 4 & 1904 \\
\addlinespace
Wave abnormality
            & LQRSV & 182 & 374 & 794 & 249 & 0 & 1599 \\
            & PRWP & 0 & 0 & 638 & 0 & 0 & 638 \\
            & QAb & 548 & 464 & 828 & 235 & 1 & 2076 \\
            & TAb & 2345 & 2306 & 4831 & 1876 & 22 & 11380 \\
            & TInv & 294 & 812 & 2720 & 157 & 5 & 3988 \\
\addlinespace
Others  
            & NSR & 18092 & 1752 & 6299 & 1826 & 922 & 28891 \\
            & Brady & 0 & 6 & 7 & 0 & 271 & 284 \\
            & PR & 296 & 0 & 1182 & 0 & 3 & 1481 \\
\bottomrule
\end{tabular*}
\label{tab:label-physionet}
\end{table}

\begin{table}[ht]
\centering
\caption{The number of labels in Alibaba dataset.} 
\small

\begin{tabular*}{\columnwidth}{@{\extracolsep{\fill}} l l l}
\hline
{Category}                   & Label                                          & {Total} \\ \hline
Arrhythmia                          
    & AF                               & 120            \\ 
    & SA                                  & 901            \\ 
    & STach                                 & 5264           \\ 
    & PAC                   & 314            \\ 
    & PVC                        & 543            \\ 
    \addlinespace

Conduction disorder                 
    & RBBB                           & 551            \\ 
    & CRBBB                  & 418            \\ 
    & 1AVB                    & 142            \\ 
    & IRBBB                & 126            \\ 
    \addlinespace
    
Axis deviation                    
    & LAD                           & 1124           \\ 
    & RAD             & 1124           \\ 
    \addlinespace

                                    
Wave abnormality                    
    & TWC                                    & 3479           \\ 
    & STC                                 & 286            \\ 
    & NSTAb                   & 64             \\
    & STTC                                       & 299            \\ 
    & HLVV                    & 414            \\ 
    \addlinespace

Others                              
    & NSR                                      & 9501           \\ 
    \bottomrule
    \end{tabular*}
\label{tab:label-Alibaba}

\end{table}

\section{Selected Hyperparameters for All Available Scaling Parameter Combinations}
\label{append:b}
The table \ref{hpo:physionet} and table \ref{hpo:Alibaba} show the performance depending on the hyperparameter optimization. \textit{F1} means the F-1 score, $K$ the kernel size, $C$ the number of channels, $D$ the depth of layers, \textit{LR} a learning rate, \textit{WD} a weight decay, \textit{Dropout} the rate of dropout, \textit{N (Aug)} the selected number of data augmentation method, \textit{M (Aug)} the intensive of data augmentation methods, and \textit{Beta} distribution of MixUP. The table is sorted in descending order by F1-score.
The search space can be found in Table \ref{tab:space}.
\newcolumntype{L}[1]{>{\raggedright\arraybackslash}p{#1}}
\small
\begin{longtable}{@{} L{1.2cm} L{0.6cm} L{0.6cm} L{0.6cm} L{2.0cm} L{2.0cm} L{1.2cm} L{1.2cm} L{1.2cm} L{1.2cm} @{}}
    \caption{Selected hyperparameters of the model in Physionet 2021 dataset.\label{hpo:physionet}} \\
    \toprule
    F1 & K & C & D & LR & WD & Dropout & N(Aug) & M(Aug) & Beta \\
    \midrule
    \endfirsthead
    
    \toprule
    F1 & K & C & D & LR & WD & Dropout & N(Aug) & M(Aug) & Beta \\
    \midrule
    \endhead
    
    \bottomrule
    \multicolumn{10}{r}{Continued on next page} \\
    \endfoot
    
    \bottomrule
    \endlastfoot
    
    
    0.6688 & 3 & 128 & 4 & 0.00260079 & 8.9745E-06 & 0.1 & 1 & 1 & 0.1 \\
    0.6646 & 3 & 64 & 2 & 0.00260079 & 8.9745E-06 & 0.1 & 1 & 1 & 0.1 \\
    0.6642 & 5 & 128 & 4 & 0.00260079 & 8.9745E-06 & 0.1 & 1 & 1 & 0.1 \\
    0.6630 & 9 & 128 & 2 & 0.00260079 & 8.9745E-06 & 0.1 & 1 & 1 & 0.1 \\
    0.6625 & 3 & 64 & 8 & 0.00117048 & 1.99539E-05 & 0 & 1 & 8 & 0.1 \\
    0.6622 & 3 & 128 & 2 & 0.000468866 & 4.48496E-06 & 0.25 & 1 & 7 & 0.2 \\
    0.6613 & 3 & 128 & 16 & 0.00117048 & 1.99539E-05 & 0 & 1 & 8 & 0.1 \\
    0.6597 & 5 & 64 & 4 & 0.00260079 & 8.9745E-06 & 0.1 & 1 & 1 & 0.1 \\
    0.6595 & 3 & 64 & 4 & 0.00260079 & 8.9745E-06 & 0.1 & 1 & 1 & 0.1 \\
    0.6585 & 5 & 128 & 2 & 0.00258996 & 1.19374E-05 & 0.15 & 0 & 7 & 0.1 \\
    0.6584 & 5 & 64 & 2 & 0.00946404 & 1.74804E-06 & 0.1 & 1 & 5 & 0 \\
    0.6583 & 3 & 128 & 8 & 0.00117048 & 1.99539E-05 & 0 & 1 & 8 & 0.1 \\
    0.6566 & 5 & 128 & 8 & 0.00260079 & 8.9745E-06 & 0.1 & 1 & 1 & 0.1 \\
    0.6547 & 3 & 32 & 2 & 0.00260079 & 8.9745E-06 & 0.1 & 1 & 1 & 0.1 \\
    0.6540 & 5 & 32 & 2 & 0.00946404 & 1.74804E-06 & 0.1 & 1 & 5 & 0 \\
    0.6537 & 9 & 64 & 2 & 0.00953107 & 2.05523E-06 & 0.05 & 0 & 8 & 0.2 \\
    0.6526 & 5 & 128 & 16 & 0.00117048 & 1.99539E-05 & 0 & 1 & 8 & 0.1 \\
    0.6523 & 3 & 64 & 16 & 0.00117048 & 1.99539E-05 & 0 & 1 & 8 & 0.1 \\
    0.6497 & 9 & 128 & 8 & 0.00117048 & 1.99539E-05 & 0 & 1 & 8 & 0.1 \\
    0.6490 & 5 & 64 & 8 & 0.00117048 & 1.99539E-05 & 0 & 1 & 8 & 0.1 \\
    0.6489 & 9 & 32 & 2 & 0.00953107 & 2.05523E-06 & 0.05 & 0 & 8 & 0.2 \\
    0.6482 & 9 & 64 & 4 & 0.00260079 & 8.9745E-06 & 0.1 & 1 & 1 & 0.1 \\
    0.6470 & 5 & 32 & 4 & 0.00260079 & 8.9745E-06 & 0.1 & 1 & 1 & 0.1 \\
    0.6469 & 9 & 128 & 16 & 0.00117048 & 1.99539E-05 & 0 & 1 & 8 & 0.1 \\
    0.6459 & 15 & 128 & 2 & 0.00260079 & 8.9745E-06 & 0.1 & 1 & 1 & 0.1 \\
    0.6450 & 9 & 128 & 4 & 0.00351696 & 1.17341E-06 & 0.05 & 2 & 3 & 0 \\
    0.6420 & 5 & 64 & 16 & 0.00117048 & 1.99539E-05 & 0 & 1 & 8 & 0.1 \\
    0.6396 & 3 & 32 & 4 & 0.00260079 & 8.9745E-06 & 0.1 & 1 & 1 & 0.1 \\
    0.6379 & 15 & 32 & 2 & 0.00260079 & 8.9745E-06 & 0.1 & 1 & 1 & 0.1 \\
    0.6376 & 15 & 64 & 2 & 0.00260079 & 8.9745E-06 & 0.1 & 1 & 1 & 0.1 \\
    0.6365 & 3 & 32 & 8 & 0.00117048 & 1.99539E-05 & 0 & 1 & 8 & 0.1 \\
    0.6360 & 9 & 32 & 4 & 0.00260079 & 8.9745E-06 & 0.1 & 1 & 1 & 0.1 \\
    0.6358 & 15 & 128 & 4 & 0.00260079 & 8.9745E-06 & 0.1 & 1 & 1 & 0.1 \\
    0.6348 & 9 & 64 & 8 & 0.00117048 & 1.99539E-05 & 0 & 1 & 8 & 0.1 \\
    0.6342 & 5 & 32 & 8 & 0.00117048 & 1.99539E-05 & 0 & 1 & 8 & 0.1 \\
    0.6336 & 15 & 64 & 4 & 0.00260079 & 8.9745E-06 & 0.1 & 1 & 1 & 0.1 \\
    0.6320 & 3 & 32 & 16 & 0.00117048 & 1.99539E-05 & 0 & 1 & 8 & 0.1 \\
    0.6302 & 9 & 32 & 8 & 0.00117048 & 1.99539E-05 & 0 & 1 & 8 & 0.1 \\
    0.6298 & 15 & 64 & 8 & 0.00117048 & 1.99539E-05 & 0 & 1 & 8 & 0.1 \\
    0.6287 & 15 & 32 & 4 & 0.00260079 & 8.9745E-06 & 0.1 & 1 & 1 & 0.1 \\
    0.6278 & 15 & 128 & 8 & 0.00117048 & 1.99539E-05 & 0 & 1 & 8 & 0.1 \\
    0.6265 & 9 & 64 & 16 & 0.00117048 & 1.99539E-05 & 0 & 1 & 8 & 0.1 \\
    0.6261 & 5 & 32 & 16 & 0.00117048 & 1.99539E-05 & 0 & 1 & 8 & 0.1 \\
    0.6250 & 15 & 32 & 8 & 0.00117048 & 1.99539E-05 & 0 & 1 & 8 & 0.1 \\
    0.6240 & 9 & 32 & 16 & 0.00117048 & 1.99539E-05 & 0 & 1 & 8 & 0.1 \\
    0.6232 & 15 & 64 & 16 & 0.00117048 & 1.99539E-05 & 0 & 1 & 8 & 0.1 \\
    0.6220 & 15 & 128 & 16 & 0.00117048 & 1.99539E-05 & 0 & 1 & 8 & 0.1 \\
    0.6205 & 15 & 32 & 16 & 0.00117048 & 1.99539E-05 & 0 & 1 & 8 & 0.1 \\
\end{longtable}

\newcolumntype{L}[1]{>{\raggedright\arraybackslash}p{#1}}
\small
\begin{longtable}{@{} L{1.2cm} L{0.6cm} L{0.6cm} L{0.6cm} L{2.0cm} L{2.0cm} L{1.2cm} L{1.2cm} L{1.2cm} L{1.2cm} @{}}
    \caption{Selected hyperparameters of the model in Alibaba dataset.\label{hpo:Alibaba}} \\
    \toprule
    F1 & K & C & D & LR & WD & Dropout & N(Aug) & M(Aug) & Beta \\
    \midrule
    \endfirsthead
    
    \toprule
    F1 & K & C & D & LR & WD & Dropout & N(Aug) & M(Aug) & Beta \\
    \midrule
    \endhead
    
    \bottomrule
    \multicolumn{10}{r}{Continued on next page} \\
    \endfoot
    
    \bottomrule
    \endlastfoot

0.5455  & 3	& 128	& 4	& 0.000597472	& 7.41711e-06	& 0.0	& 2 &	7	& 0.1 \\
0.5397	& 5	& 128	& 4	& 0.00107348	& 1.17844e-06	& 0.1	& 2 &	1	& 0.1 \\
0.5366	& 3	& 128	& 8	& 0.00107348	& 1.17844e-06	& 0.1	& 2 &	1	& 0.1 \\
0.5345	& 3	& 16	& 4	& 0.00107348	& 1.17844e-06	& 0.1	& 2 &	1	& 0.1 \\
0.533	& 9	& 128	& 4	& 0.00677992	& 3.99025e-06	& 0.0	& 2 &	3	& 0.0 \\
0.5247	& 5	& 128	& 2	& 0.00677992	& 3.99025e-06	& 0.0	& 2 &	3	& 0.0 \\
0.5241	& 5	& 32	& 2	& 0.00107348	& 1.17844e-06	& 0.1	& 2 &	1	& 0.1 \\
0.5236	& 5	& 64	& 2	& 0.00677992	& 3.99025e-06	& 0.0	& 2 &	3	& 0.0 \\
0.5227	& 3	& 64	& 2	& 0.00107348	& 1.17844e-06	& 0.1	& 2 &	1	& 0.1 \\
0.5219	& 3	& 32	& 4	& 0.00677992	& 3.99025e-06	& 0.0	& 2 &	3	& 0.0 \\
0.5204	& 3	& 32	& 8	& 0.00161031	& 1.39832e-05	& 0.0	& 0 &	6	& 0.2 \\
0.5201	& 9	& 32	& 2	& 0.00932976	& 3.59464e-06	& 0.1	& 2 &	3	& 0.0 \\
0.52    & 3	& 64	& 4	& 0.00107348	& 1.17844e-06	& 0.1	& 2 &	1	& 0.1 \\
0.5186	& 3	& 64	& 8	& 0.00161031	& 1.39832e-05	& 0.0	& 0 &	6	& 0.2 \\
0.5181	& 5	& 64	& 4	& 0.00677992	& 3.99025e-06	& 0.0	& 2 &	3	& 0.0 \\
0.5178	& 3	& 128	& 16& 0.000597472	& 7.41711e-06	& 0.0	& 2 &	7	& 0.1 \\
0.5166	& 5	& 128	& 16& 0.00161031	& 1.39832e-05	& 0.0	& 0 &	6	& 0.2 \\
0.5164	& 3	& 128	& 2	& 0.00107348	& 1.17844e-06	& 0.1	& 2 &	1	& 0.1 \\
0.5164	& 3	& 16	& 8	& 0.000597472	& 7.41711e-06	& 0.0	& 2 &	7	& 0.1 \\
0.5153	& 3	& 32	& 2	& 0.000806845	& 6.48126e-06	& 0.2	& 0	&   10	& 0.2 \\
0.5147	& 15& 128	& 4	& 0.000979419	& 8.11599e-06	& 0.1	& 0 &	5	& 0.2 \\
0.5132	& 9	& 128	& 8	& 0.000597472	& 7.41711e-06	& 0.0	& 2 &	7	& 0.1 \\
0.513	& 5	& 64	& 8	& 0.000597472	& 7.41711e-06	& 0.0	& 2 &	7	& 0.1 \\
0.5098	& 5	& 128	& 8	& 0.00677992	& 3.99025e-06	& 0.0	& 2 &	3	& 0.0 \\
0.5096	& 9	& 64	& 8	& 0.00677992	& 3.99025e-06	& 0.0	& 2 &	3	& 0.0 \\
0.5089	& 9	& 64	& 4	& 0.000490151	& 2.57339e-06	& 0.2	& 0 &	7	& 0.2 \\
0.5076	& 3	& 32	& 16& 0.000597472	& 7.41711e-06	& 0.0	& 2 &	7	& 0.1 \\
0.5027	& 3	& 16	& 2	& 0.00677992	& 3.99025e-06	& 0.0	& 2 &	3	& 0.0 \\
0.5023	& 9	& 128	& 2	& 0.00107348	& 1.17844e-06	& 0.1	& 2 &	1	& 0.1 \\
0.5021	& 15& 16	& 2	& 0.00697859	& 1.78065e-05	& 0.0	& 1 &	4	& 0.1 \\
0.5003	& 9	& 64	& 2	& 0.00107348	& 1.17844e-06	& 0.1	& 2 &	1	& 0.1 \\
0.4993	& 9	& 32	& 8	& 0.0078945	    & 1.10935e-05	& 0.0	& 1 &	7	& 0.1 \\
0.4992	& 5	& 32	& 16& 0.00161031	& 1.39832e-05	& 0.0	& 0 &	6	& 0.2 \\
0.4987	& 5	& 32	& 8	& 0.000597472	& 7.41711e-06	& 0.0	& 2 &	7	& 0.1 \\
0.4979	& 15& 128   & 2 & 0.00107348	& 1.17844e-06	& 0.1	& 2 &	1	& 0.1 \\
0.4968	& 5	& 32	& 4	& 0.00677992	& 3.99025e-06	& 0.0	& 2 &	3	& 0.0 \\
0.494	& 5	& 16	& 8	& 0.00677992	& 3.99025e-06	& 0.0	& 2 &	3	& 0.0 \\
0.4936	& 5	& 64	& 16& 0.00541542	& 6.89599e-06	& 0.0	& 0 &	4	& 0.0 \\
0.4925	& 3	& 64	& 16& 0.000597472	& 7.41711e-06	& 0.0	& 2 &	7	& 0.1 \\
0.4924	& 15& 64	& 16& 0.000597472	& 7.41711e-06	& 0.0	& 2 &	7	& 0.1 \\
0.4917	& 3	& 16	& 16& 0.00161031	& 1.39832e-05	& 0.0	& 0 &	6	& 0.2 \\
0.4863	& 9	& 64	& 16& 0.00677992	& 3.99025e-06	& 0.0	& 2 &	3	& 0.0 \\
0.4858	& 9	& 16	& 2	& 0.00677992	& 3.99025e-06	& 0.0	& 2 &	3	& 0.0 \\
0.4847	& 15& 64	& 8 & 0.00161031	& 1.39832e-05	& 0.0	& 0 &	6	& 0.2 \\
0.4835	& 15& 32	& 2 & 0.00161031	& 1.39832e-05	& 0.0	& 0 &	6	& 0.2 \\
0.4828	& 15& 128	& 8 & 0.000597472	& 7.41711e-06	& 0.0	& 2 &	7	& 0.1 \\
0.4816	& 15& 32	& 4 & 0.00677992	& 3.99025e-06	& 0.0	& 2 &	3	& 0.0 \\
0.4809	& 5	& 16	& 2	& 0.00677992	& 3.99025e-06	& 0.0	& 2 &	3	& 0.0 \\
0.4808	& 15& 64	& 2	& 0.00161031	& 1.39832e-05	& 0.0	& 0 &	6	& 0.2 \\
0.4797	& 9	& 128	& 16& 0.000105932	& 7.98499e-05	& 0.0	& 1 &	2	& 0.1 \\
0.479	& 9	& 32	& 4	& 0.00107348	& 1.17844e-06	& 0.1	& 2 &	1	& 0.1 \\
0.4747	& 15& 128	& 16& 0.00697859	& 1.78065e-05	& 0.0	& 1 &	4	& 0.1 \\
0.4709	& 9	& 32	& 16& 0.000597472	& 7.41711e-06	& 0.0	& 2 &	7	& 0.1 \\
0.4702	& 5	& 16	& 16& 0.000597472	& 7.41711e-06	& 0.0	& 2 &	7	& 0.1 \\
0.4698	& 5	& 16	& 4	& 0.00697859	& 1.78065e-05	& 0.0	& 1 &	4	& 0.1 \\
0.4691	& 15& 16	& 8	& 0.00677992	& 3.99025e-06	& 0.0	& 2 &	3	& 0.0 \\
0.4688	& 15& 32	& 8	& 0.00697859	& 1.78065e-05	& 0.0	& 1 &	4	& 0.1 \\
0.4682	& 9	& 16	& 4	& 0.000979419	& 8.11599e-06	& 0.1	& 0 &	5	& 0.2 \\
0.4681	& 15& 32	& 16& 0.00161031	& 1.39832e-05	& 0.0	& 0 &	6	& 0.2 \\
0.4633	& 9	& 16	& 16& 0.000597472	& 7.41711e-06	& 0.0	& 2 &	7	& 0.1 \\
0.4561	& 9	& 16	& 8	& 0.00677992	& 3.99025e-06	& 0.0	& 2 &	3	& 0.0 \\
0.4511	& 15& 64	& 4	& 0.000297885	& 2.00125e-06	& 0.1	& 1 &	6	& 0.0 \\
0.4505	& 15& 16	& 16& 0.00677992	& 3.99025e-06	& 0.0	& 2 &	3	& 0.0 \\
0.4462	& 15& 16	& 4	& 0.00107348	& 1.17844e-06	& 0.1	& 2 &	1	& 0.1 \\

\end{longtable}

\large

\section{Selected Hyperparameters Depending on the Range of the Search Space }
\label{append:c}
The table \ref{hpo:physionet2021-optimal} and \ref{hpo:Alibaba-optimal} show the hyperparameters of the selected model provided in Table~\ref{tab:hpo-perf}. It is important to note that suboptimal scaling factors are selected when the search space is broad.

\newcolumntype{L}[1]{>{\raggedright\arraybackslash}p{#1}}
\small
\begin{longtable}{@{} L{1.2cm} L{0.6cm} L{0.6cm} L{0.6cm} L{2.0cm} L{2.0cm} L{1.2cm} L{1.2cm} L{1.2cm} L{1.2cm} @{}}

    \caption{Selected hyperparameters of the model in Physionet 2021 dataset. \label{hpo:physionet2021-optimal}}\\
    \toprule
     & K & C & D & LR & WD & Dropout & N(Aug) & M(Aug) & Mixup \\
    \midrule
    \endfirsthead
    \toprule
     & K & C & D & LR & WD & Dropout & N(Aug) & M(Aug) & beta \\
    \midrule
    \endhead
    \bottomrule
    \multicolumn{10}{r}{Continued on next page} \\
    \endfoot
    \bottomrule
    \endlastfoot
    Large & 9 & 128 & 4 & 0.000458882 & 4.99297E-06	 & 0.2 & 2 & 9 & 0.0 \\
    Medium & 5 & 128 & 2 & 0.00339825 & 6.13601E-06 & 0.1 & 2 & 1 & 0.1 \\
    Optimal & 3 & 128 & 4 & 0.00260079 & 8.9745E-06 & 0.1 & 1 & 1 & 0.1 \\
\end{longtable}

\newcolumntype{L}[1]{>{\raggedright\arraybackslash}p{#1}}
\small
\begin{longtable}{@{} L{1.2cm} L{0.6cm} L{0.6cm} L{0.6cm} L{2.0cm} L{2.0cm} L{1.2cm} L{1.2cm} L{1.2cm} L{1.2cm} @{}}

    \caption{Selected hyperparameters of the model in Alibaba dataset. \label{hpo:Alibaba-optimal}}\\
    \toprule
     & K & C & D & LR & WD & Dropout & N(Aug) & M(Aug) & Mixup \\
    \midrule
    \endfirsthead
    \toprule
     & K & C & D & LR & WD & Dropout & N(Aug) & M(Aug) & beta \\
    \midrule
    \endhead
    \bottomrule
    \multicolumn{10}{r}{Continued on next page} \\
    \endfoot
    \bottomrule
    \endlastfoot
    Large & 0 & 0 & 0 & 0.0 & 0.0E-06	 & 0.0 & 0 & 0 & 0.0 \\
    Medium & 0 & 0 & 0 & 0.0 & 0.0E-06 & 0.0 & 0 & 0 & 0.0 \\
    Optimal & 0 & 0 & 0 & 0.0 & 0.0E-06 & 0.0 & 0 & 0 & 0.0 \\
\end{longtable}

\large
\section{Classification Performance on Labels}
\label{append:d}
\begin{figure}[ht]
    \vspace{-5mm}
        \includegraphics[width=0.5\textwidth]{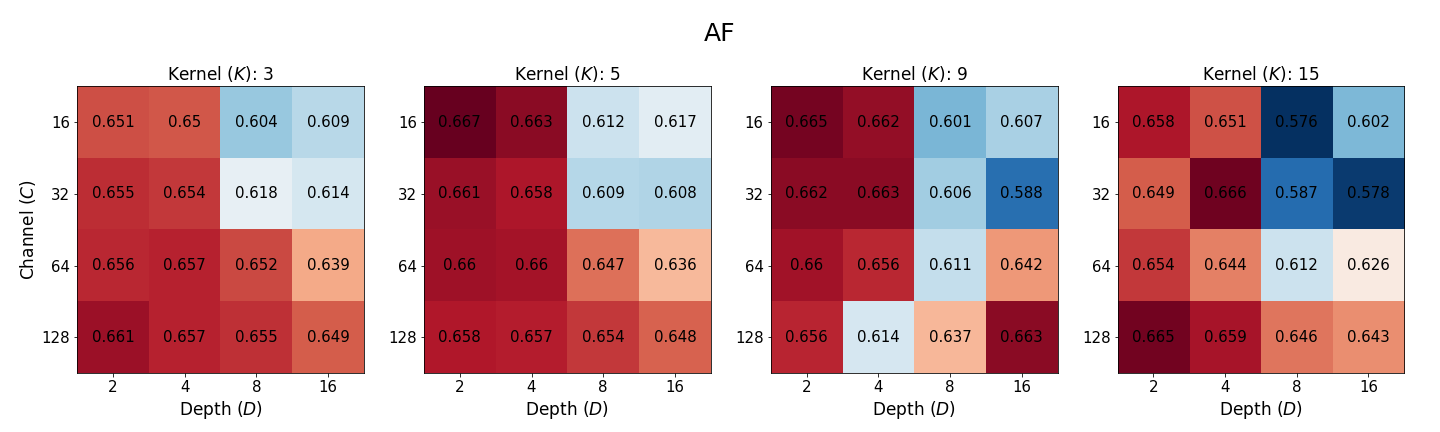}
        \includegraphics[width=0.5\textwidth]{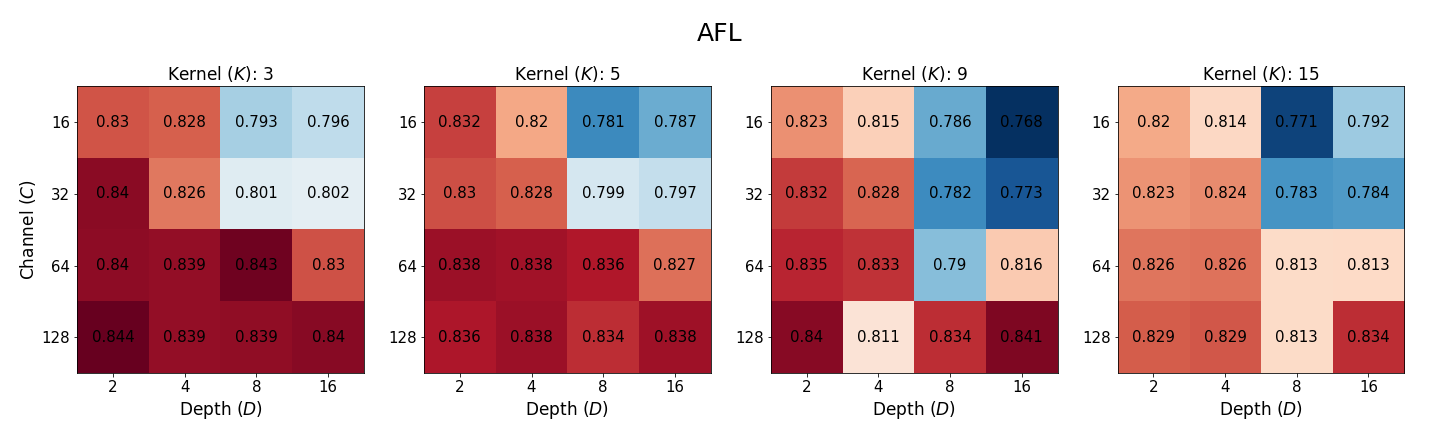}
        \includegraphics[width=0.5\textwidth]{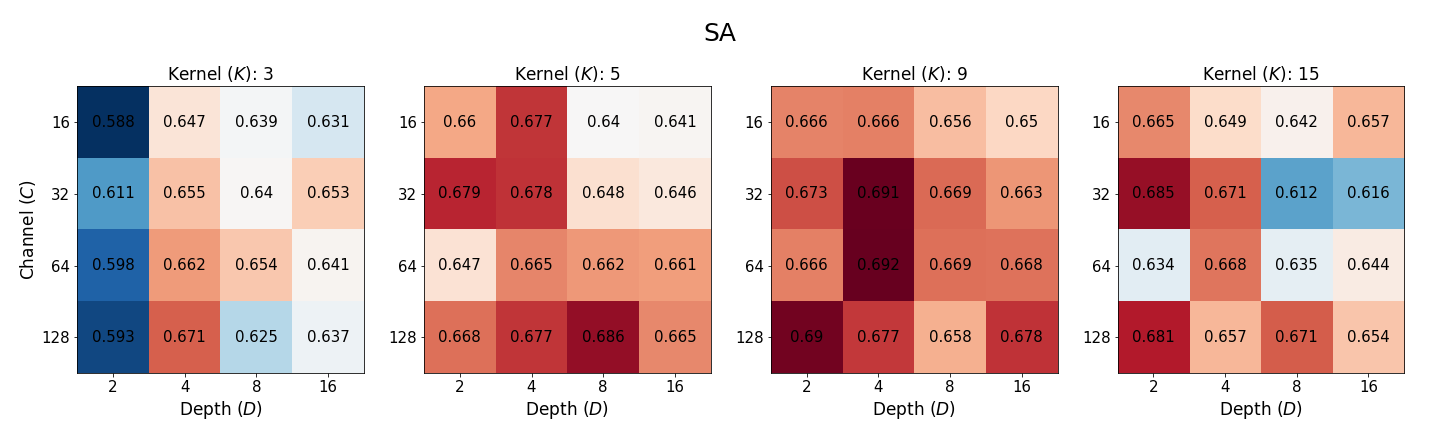}
        \includegraphics[width=0.5\textwidth]{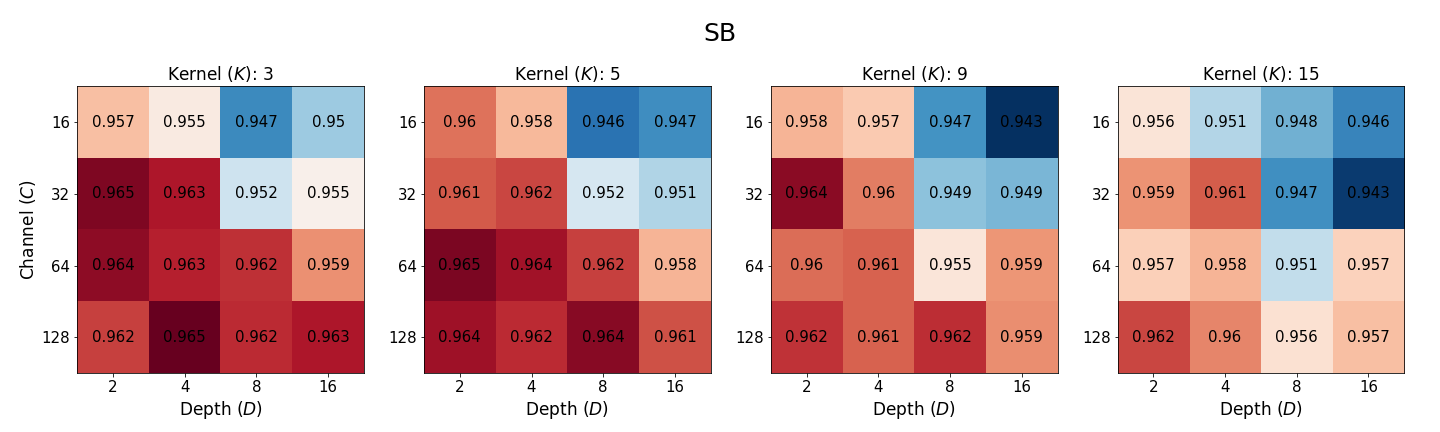}
        \includegraphics[width=0.5\textwidth]{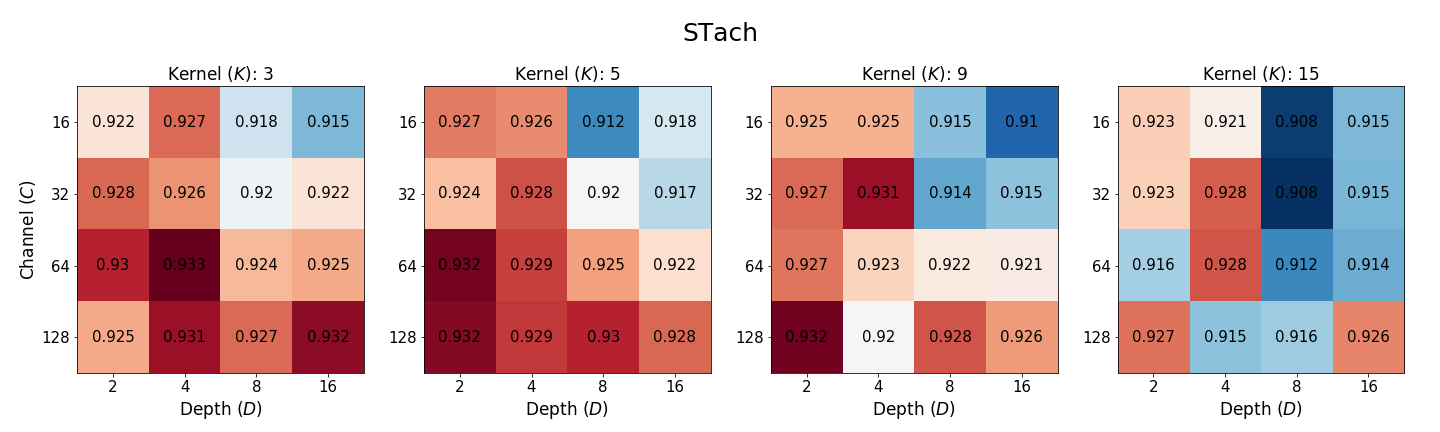}
        \includegraphics[width=0.5\textwidth]{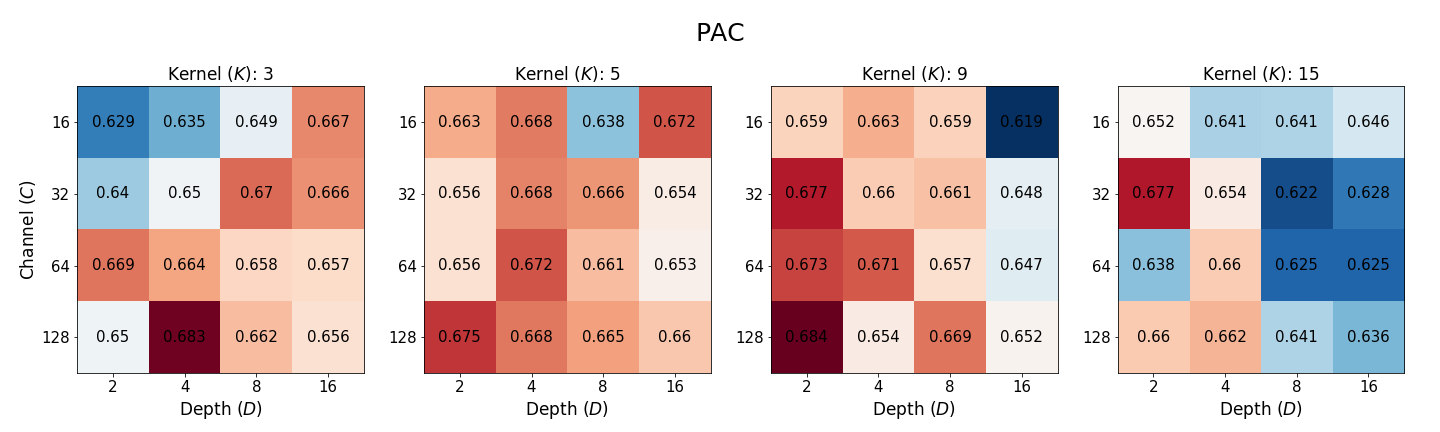}
        \includegraphics[width=0.5\textwidth]{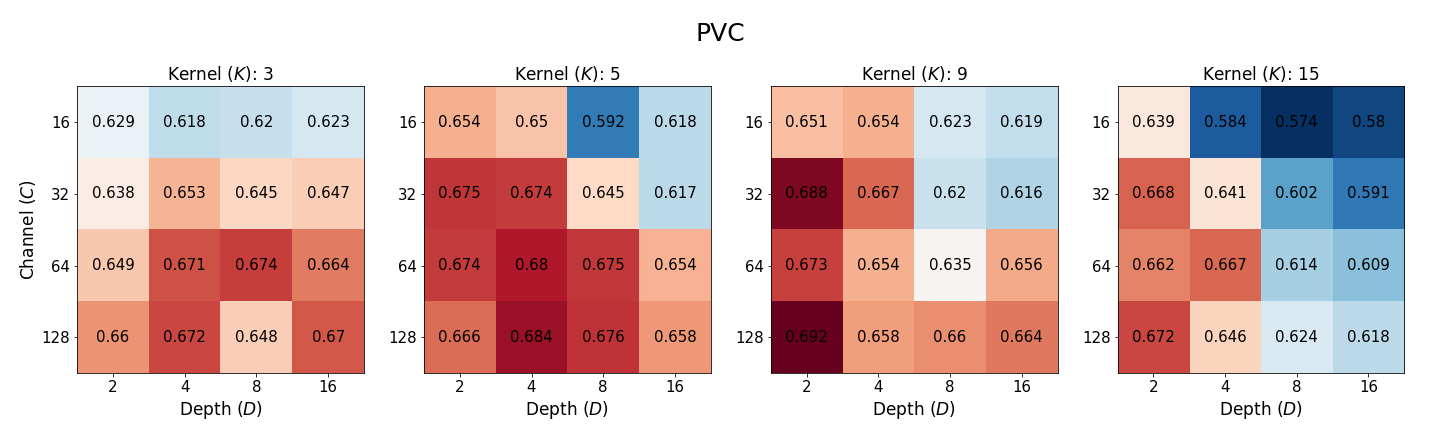}
\caption{The classification performance of networks with different layer depth, number of channels, and kernel size on arrhythmia labels in Physionet 2021 dataset.}
\label{performance-physionet-arrhythmia}
\vspace{-5mm}
\end{figure}

\begin{figure}[ht]
    \vspace{15mm}
        \includegraphics[width=0.5\textwidth]{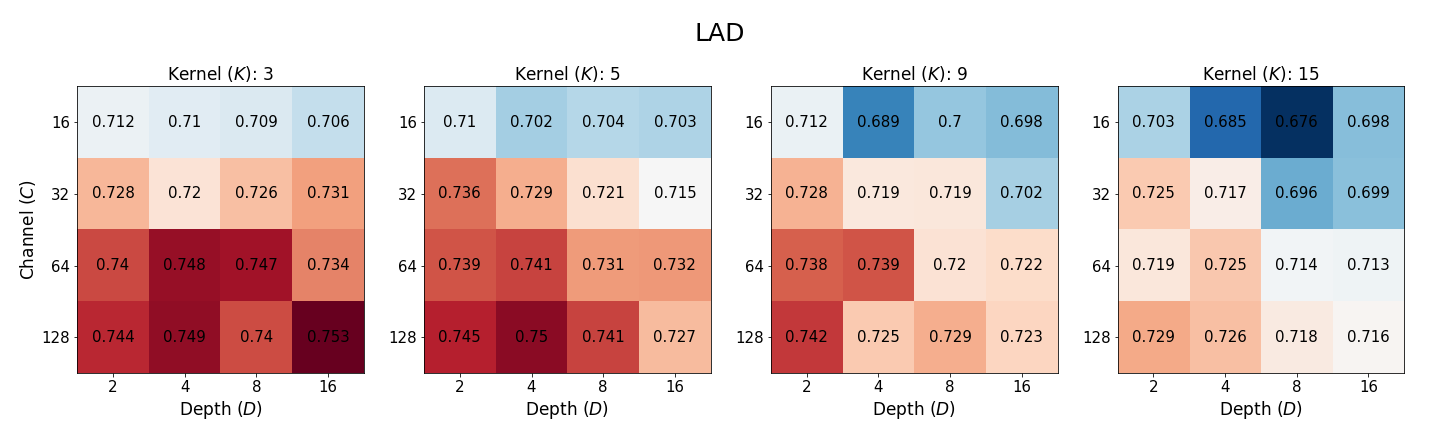}
        \includegraphics[width=0.5\textwidth]{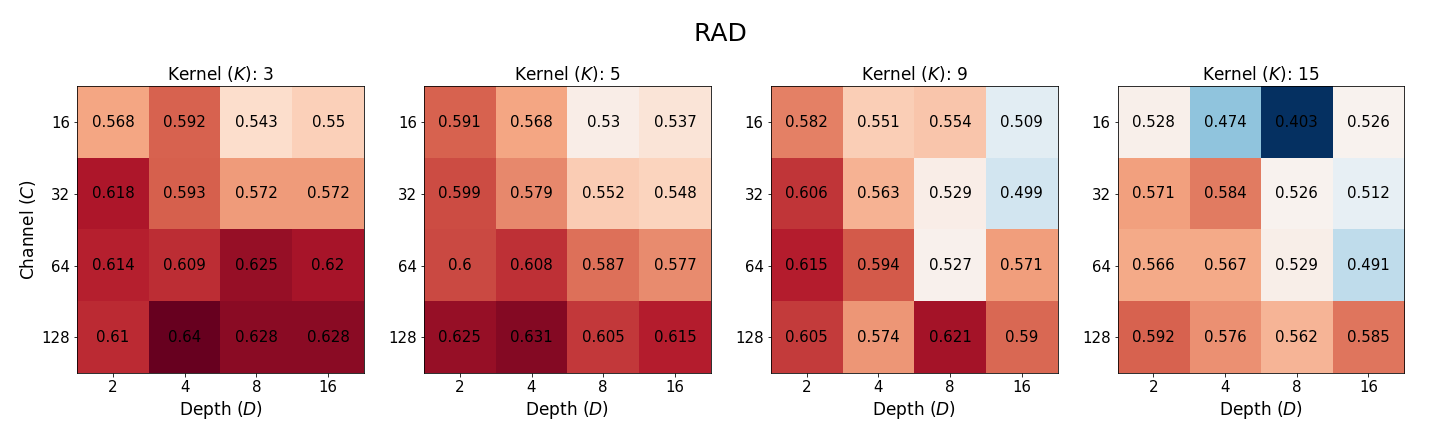}
    \caption{The classification performance of networks with different layer depth, number of channels, and kernel size on axis deviations labels in Physionet 2021 dataset.}
    \label{performance-physionet-axis}

\end{figure}

\begin{figure}[h]
        \includegraphics[width=0.5\textwidth]{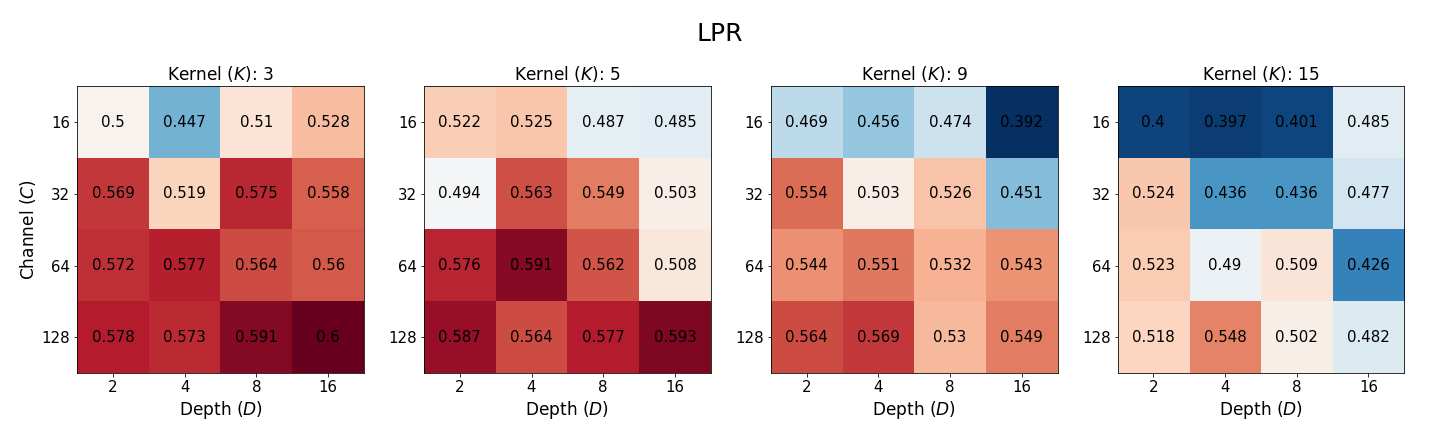}
        \includegraphics[width=0.5\textwidth]{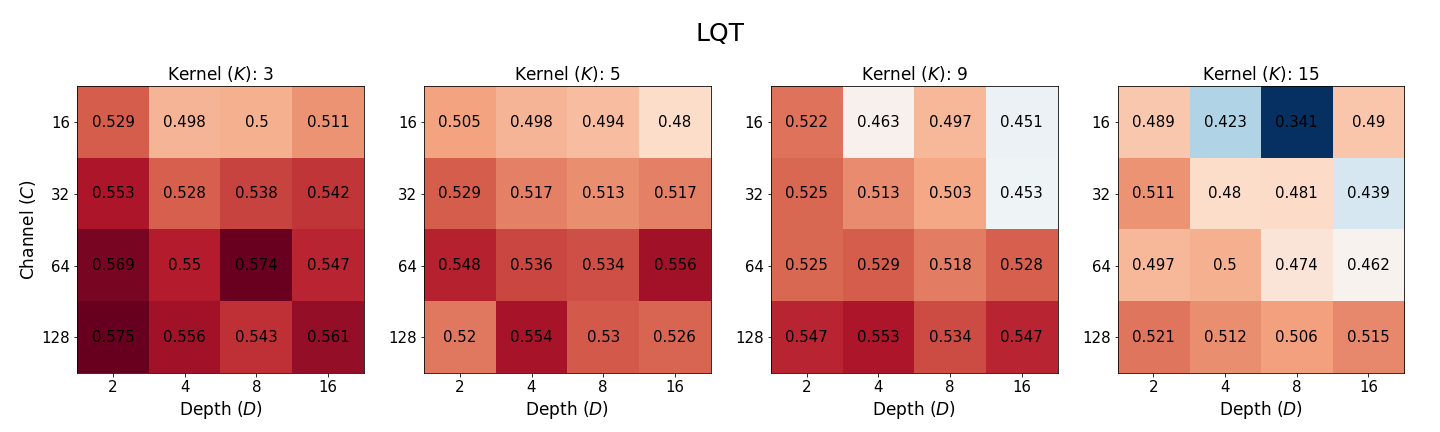}
    \caption{The classification performance of networks with different layer depth, number of channels, and kernel size on prolonged intervals labels in Physionet 2021 dataset.}
    \label{performance-physionet-interval}
\end{figure}

\begin{figure}[hb]
        \includegraphics[width=0.5\textwidth]{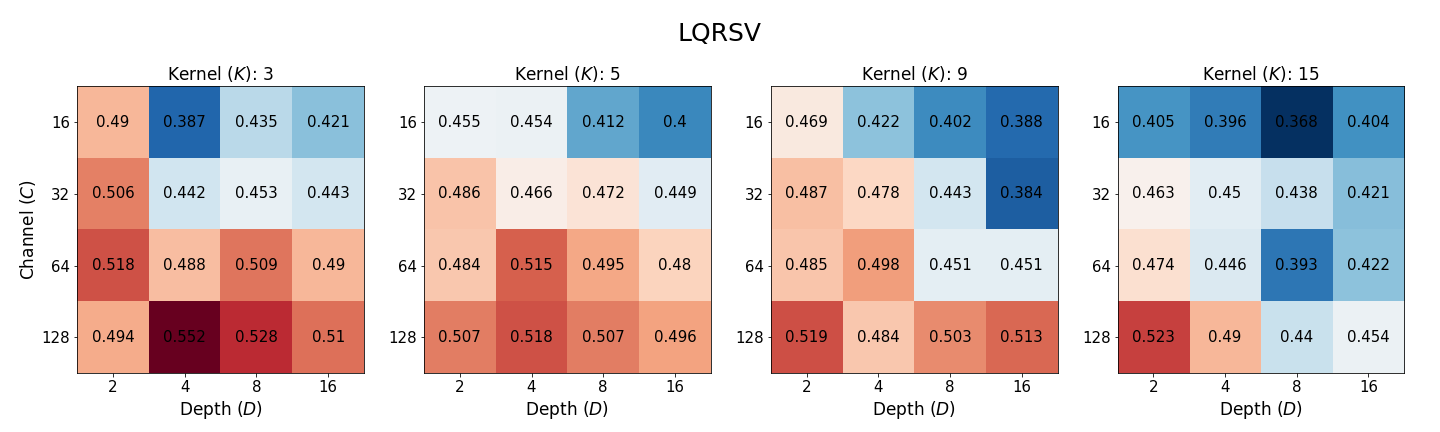}
        \includegraphics[width=0.5\textwidth]{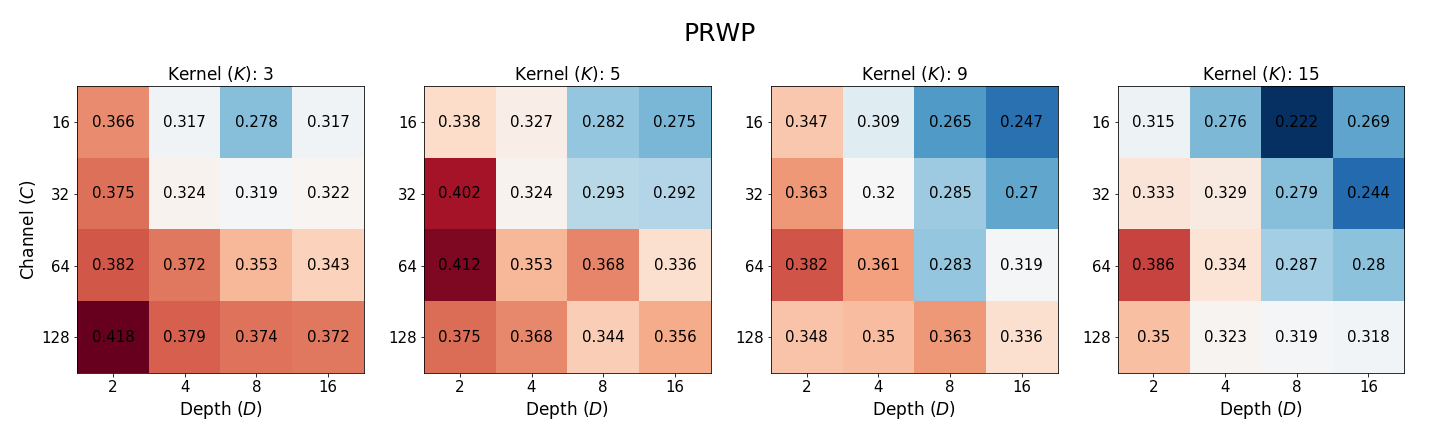}
        \includegraphics[width=0.5\textwidth]{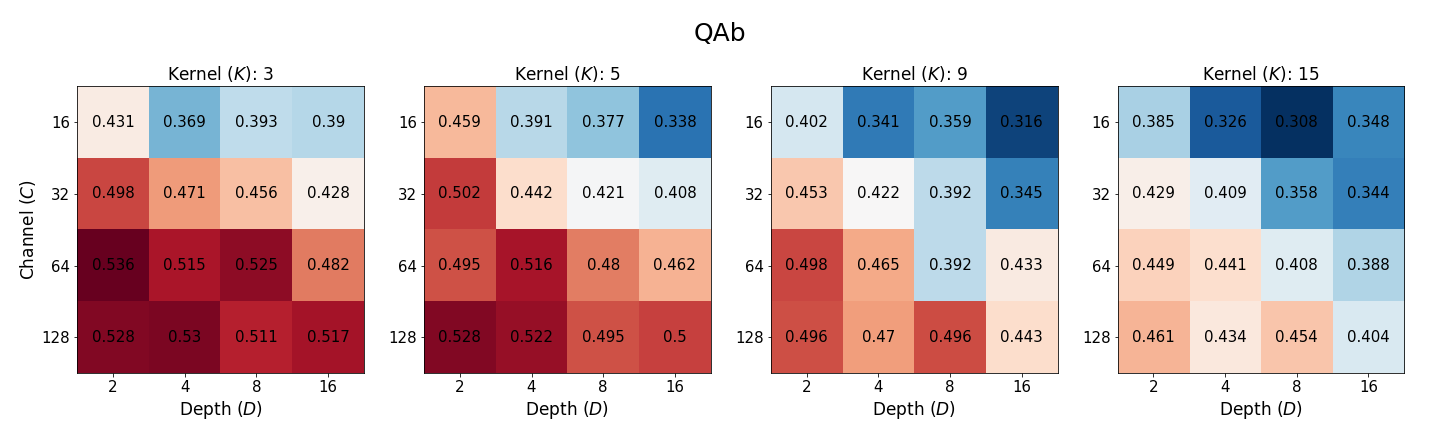}
        \includegraphics[width=0.5\textwidth]{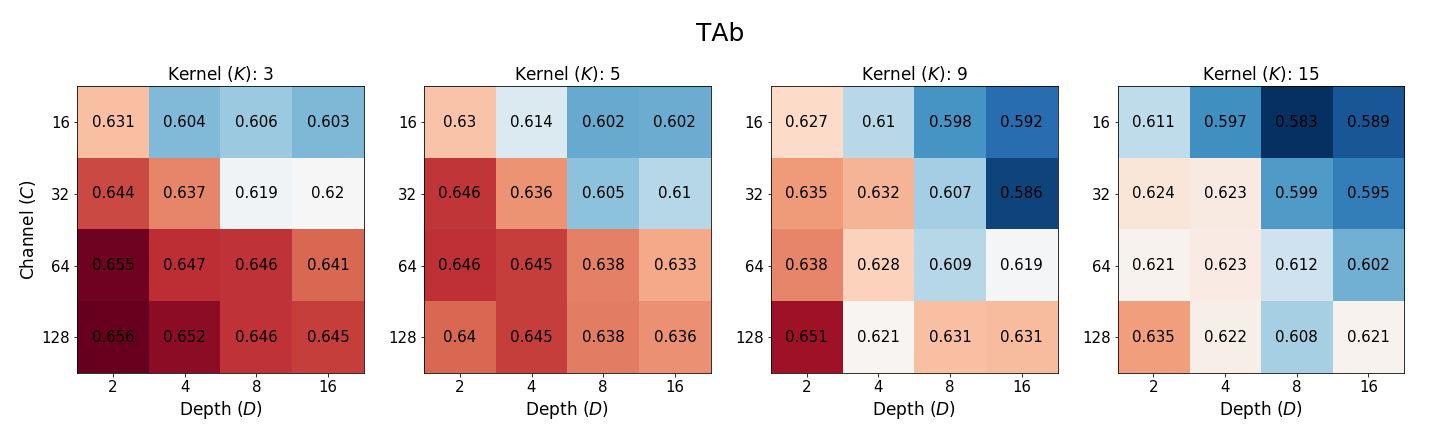}
        \includegraphics[width=0.5\textwidth]{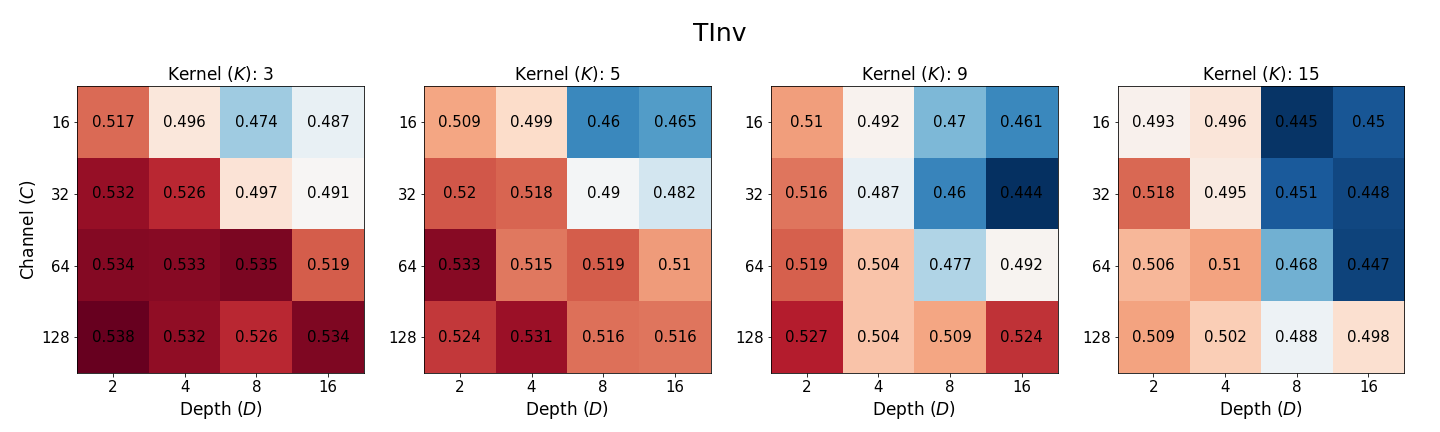}
    \caption{The classification performance of networks with different layer depth ($D$), number of channels ($C$), and kernel size ($K$) on wave abnormalities labels in Physionet 2021 dataset.
    \\
    \\
    \\
    \\
    \\
    \\
    \\}
    \label{performance-physionet-wave}

\end{figure}

\begin{figure}[ht]

        \includegraphics[width=0.5\textwidth]{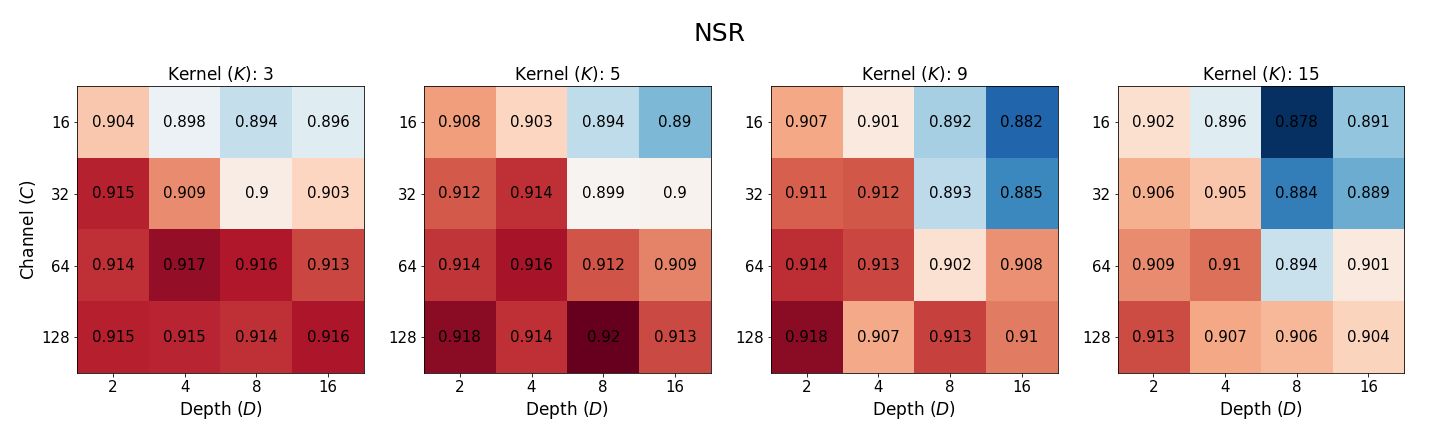}
        \includegraphics[width=0.5\textwidth]{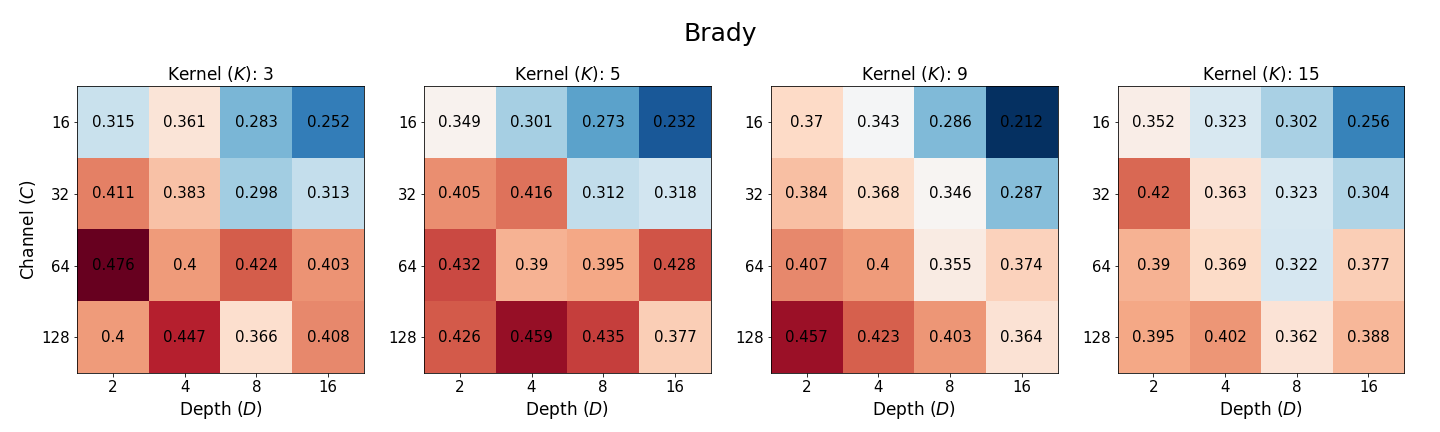}
        \includegraphics[width=0.5\textwidth]{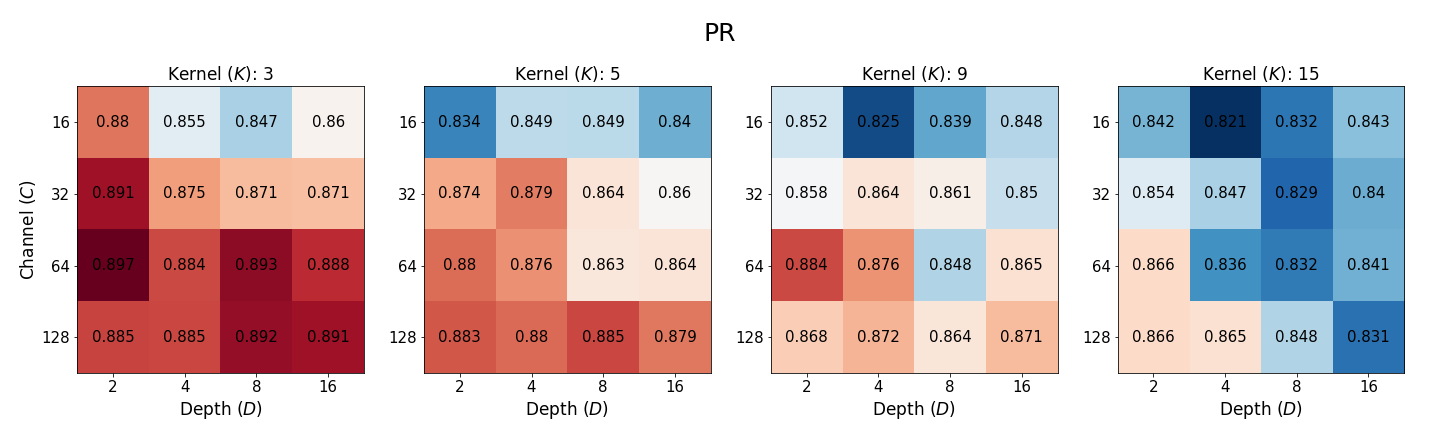}
    \caption{The classification performance of networks with different layer depth, number of channels, and kernel size on others labels in Physionet 2021 dataset.}
    \label{performance-physionet-others}

\end{figure}

\begin{figure}[h]
        \includegraphics[width=0.5\textwidth]{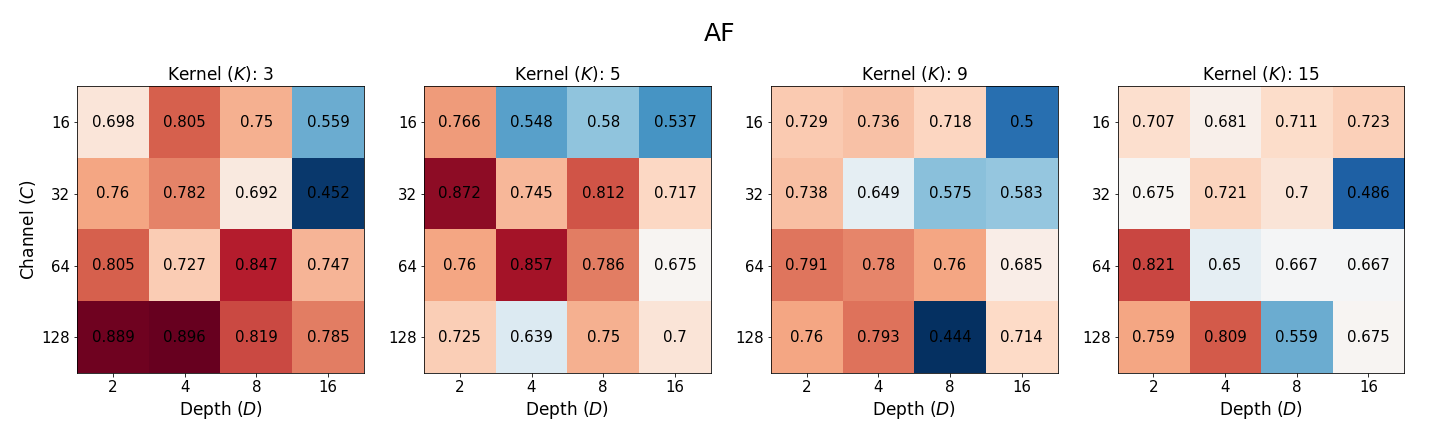}
        \includegraphics[width=0.5\textwidth]{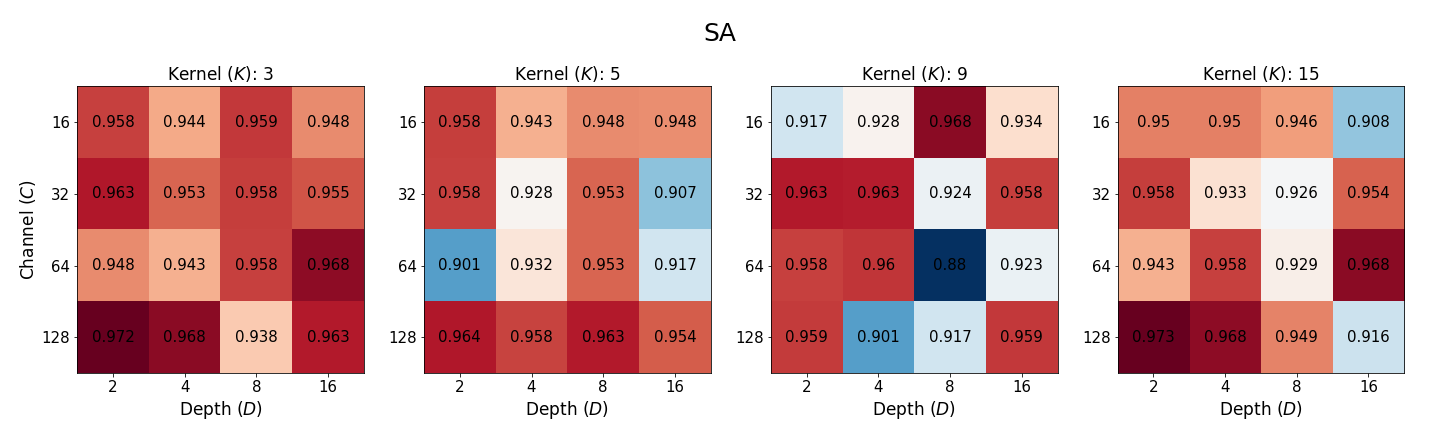}
        \includegraphics[width=0.5\textwidth]{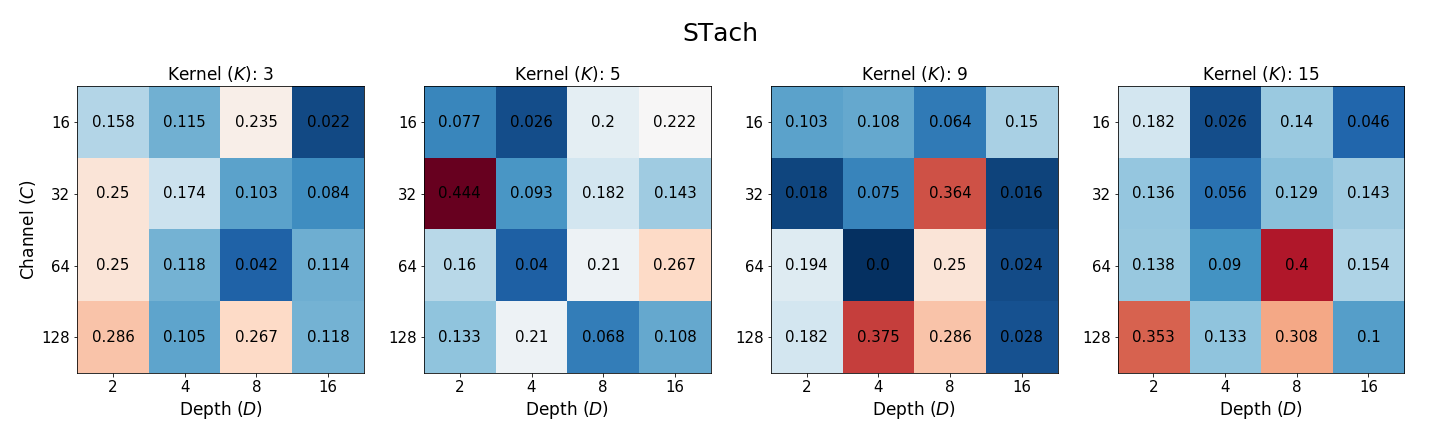}
        \includegraphics[width=0.5\textwidth]{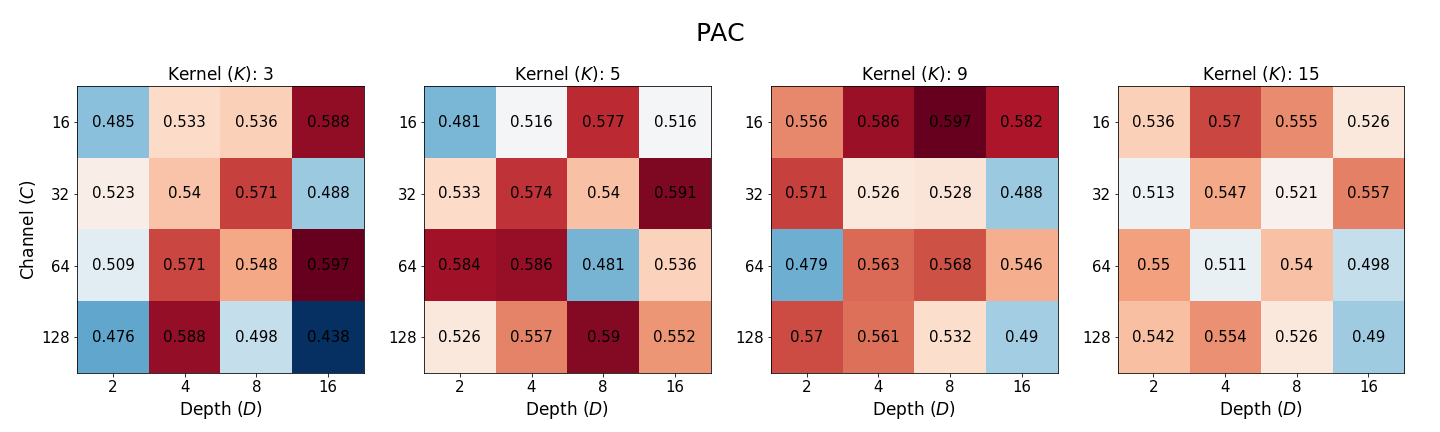}
        \includegraphics[width=0.5\textwidth]{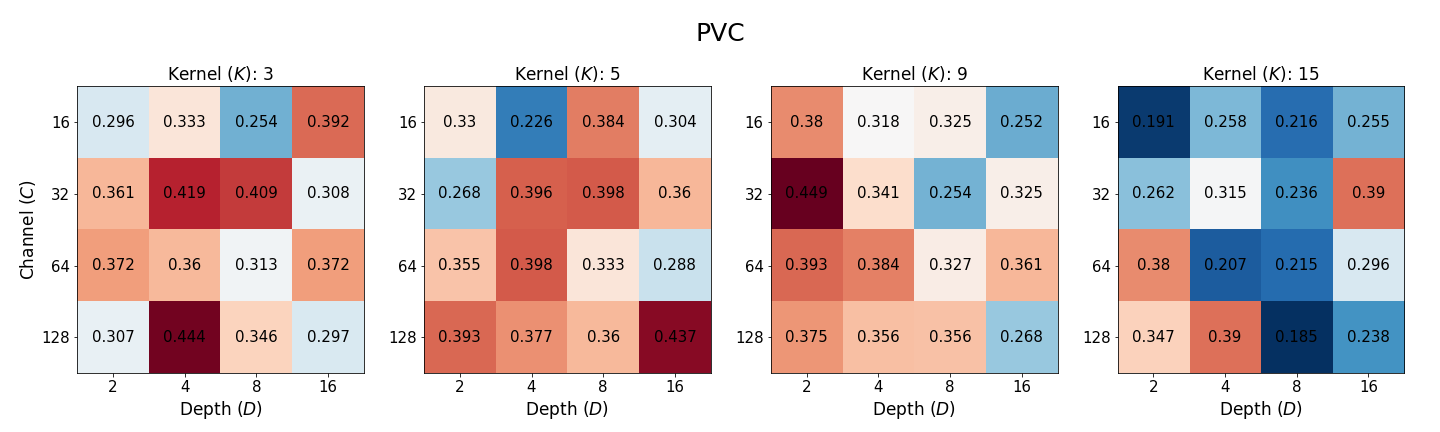}
    \caption{The classification performance of networks with different layer depth, number of channels, and kernel size on arrhythmia labels in Alibaba dataset.}
    \label{performance-Alibaba-arrhythmia}

\end{figure}

\begin{figure}[ht]
\vspace{20mm}
        \includegraphics[width=0.5\textwidth]{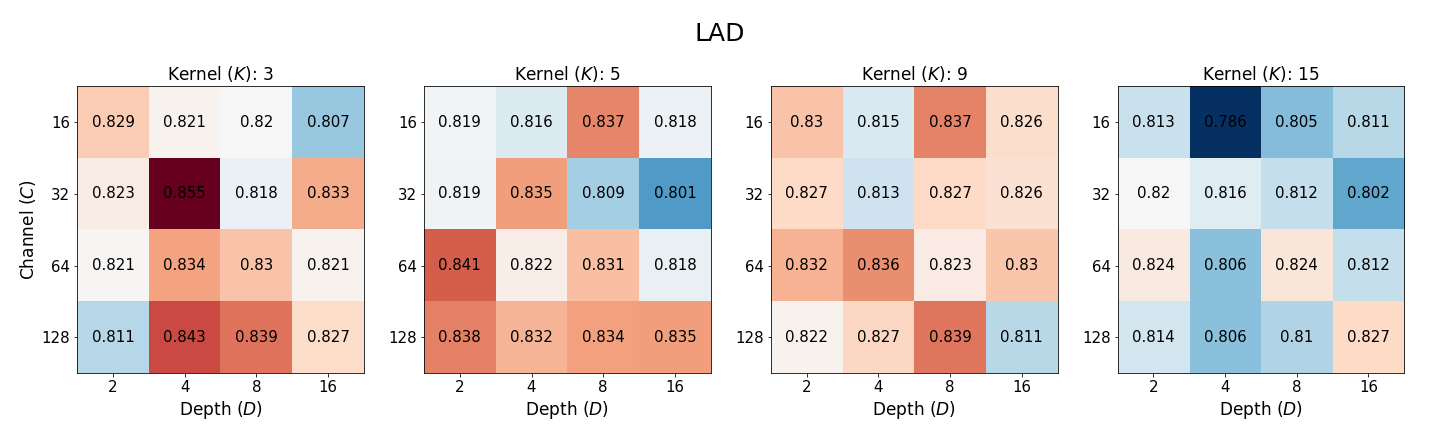}
        \includegraphics[width=0.5\textwidth]{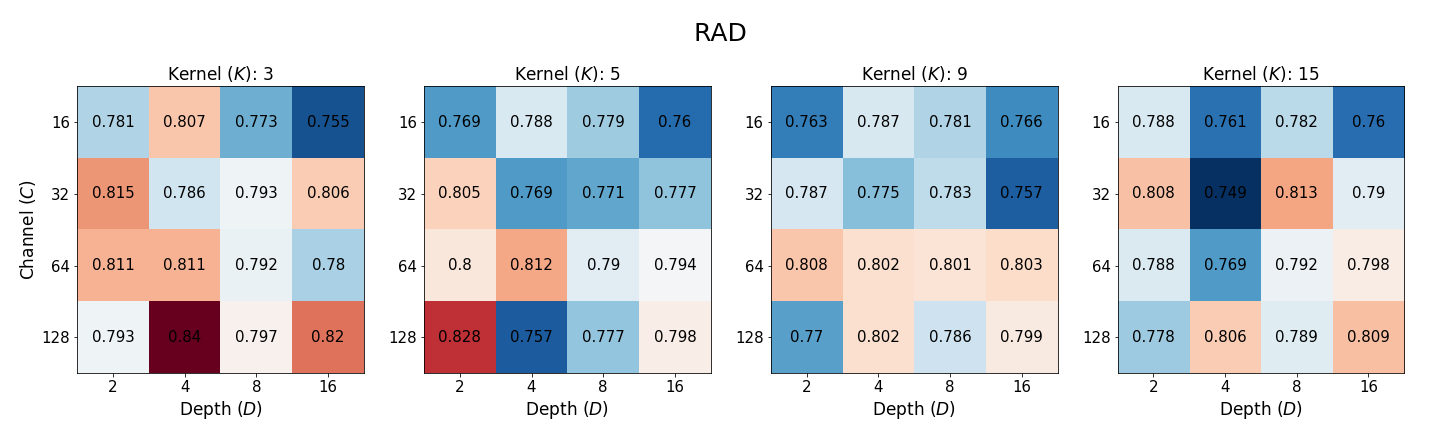}
    \caption{The classification performance of networks with different layer depth, number of channels, and kernel size on axis deviations labels in Alibaba dataset.}
    \label{performance-Alibaba-axis}

\end{figure}

\begin{figure}[h]
        \includegraphics[width=0.5\textwidth]{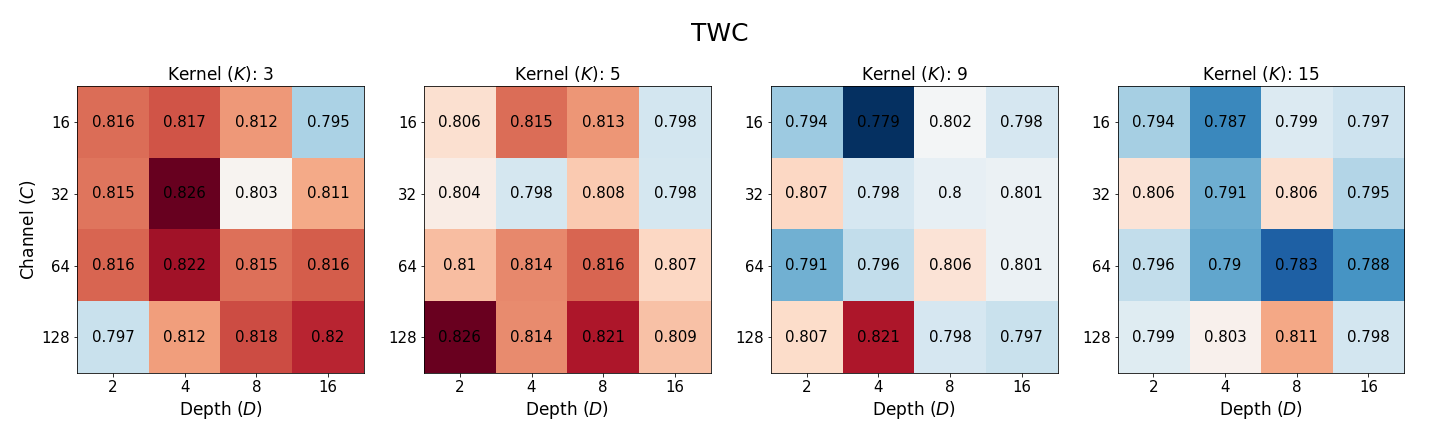}
        \includegraphics[width=0.5\textwidth]{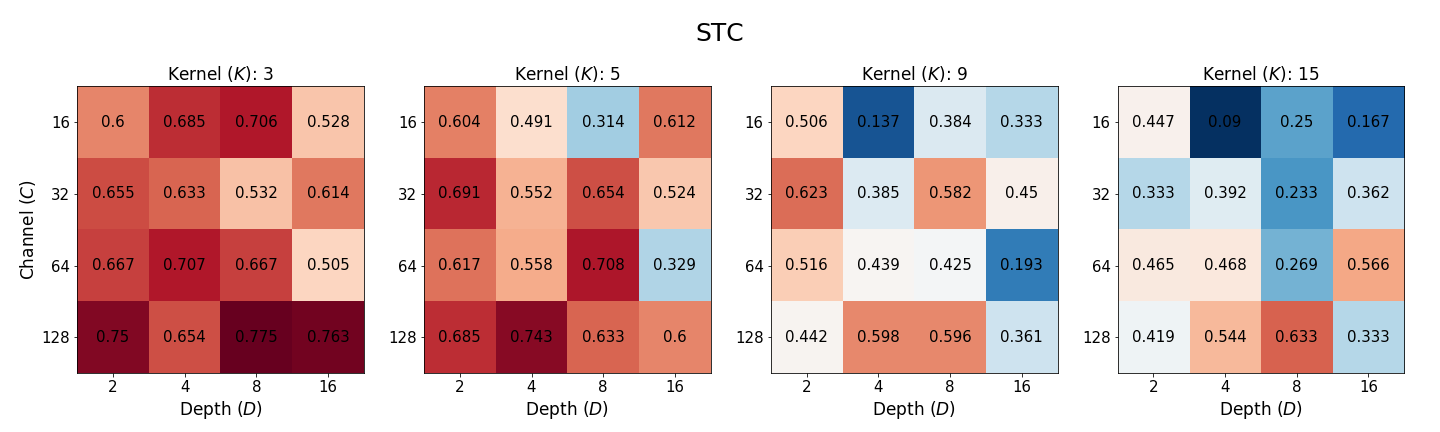}
        \includegraphics[width=0.5\textwidth]{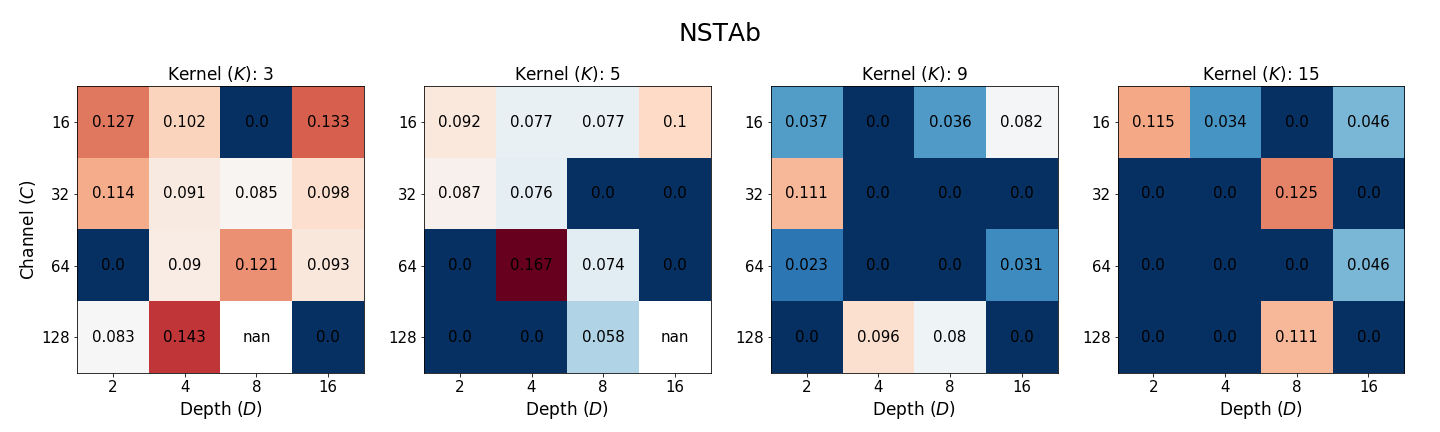}
        \includegraphics[width=0.5\textwidth]{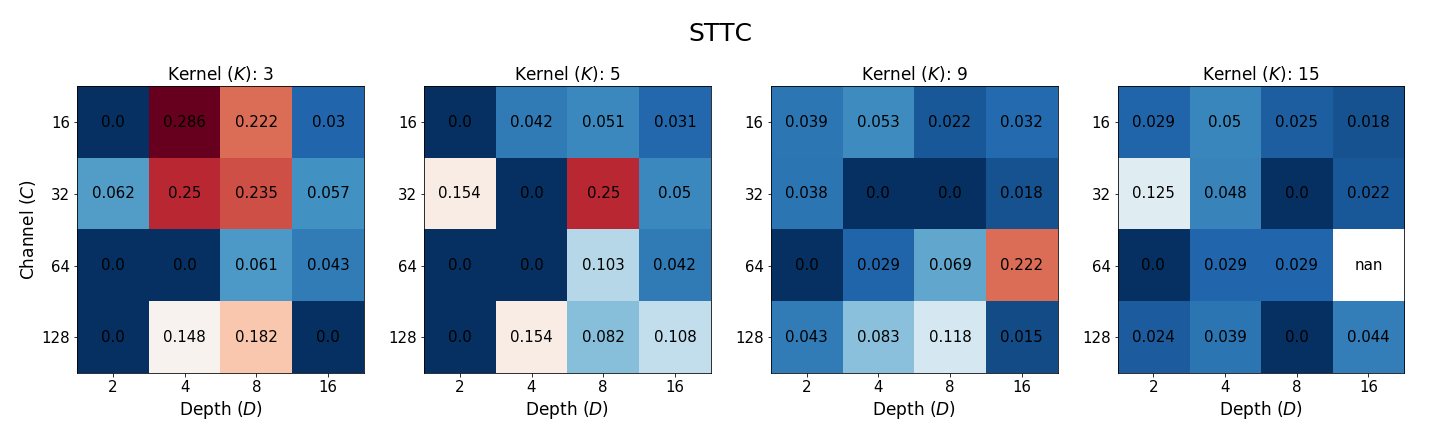}
        \includegraphics[width=0.5\textwidth]{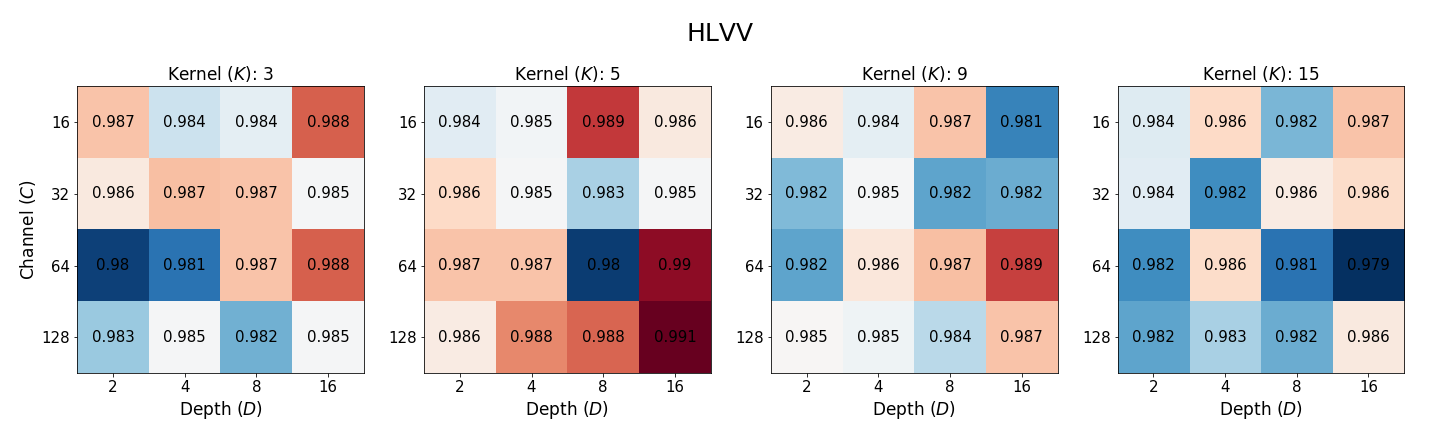}
    \caption{The classification performance of networks with different layer depth ($D$), number of channels ($C$), and kernel size ($K$) on wave abnormalities labels in Alibaba dataset.}
    \label{performance-Alibaba-wave}

\end{figure}

\begin{figure}[h]
        \includegraphics[width=0.5\textwidth]{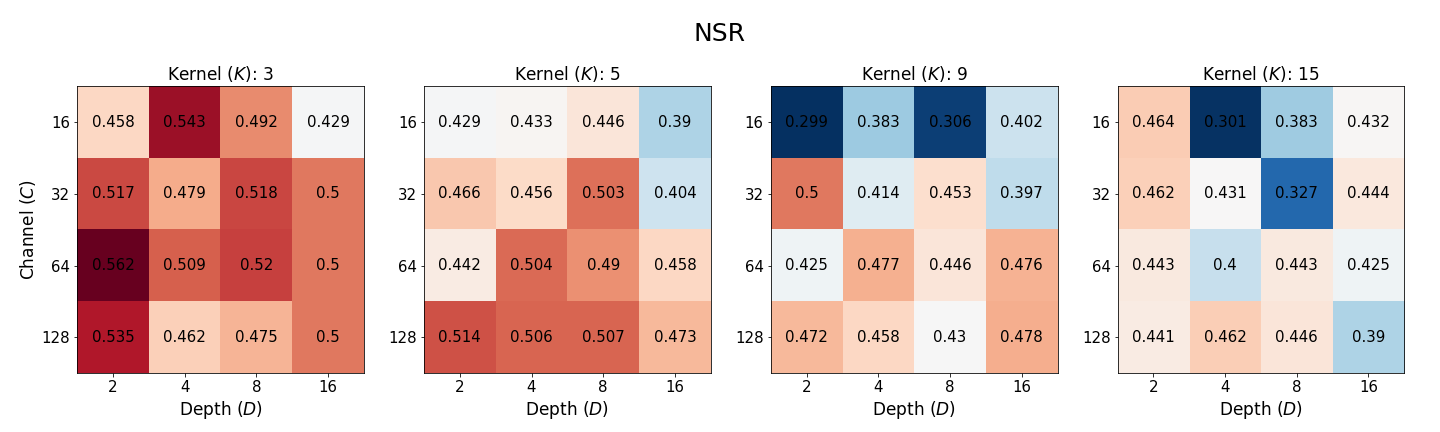}
    \caption{The classification performance of networks with different layer depth, number of channels, and kernel size on others labels in Alibaba dataset.
    \\
    \\
    \\
    \\
    \\}
    \label{performance-Alibaba-others}

\end{figure}

Figure \ref{performance-physionet-arrhythmia}, \ref{performance-physionet-axis}, \ref{performance-physionet-interval}, \ref{performance-physionet-wave}, \ref{performance-physionet-others} illustrate the classification performance across 26 distinct labels of Physionet 2021 dataset, and Figure \ref{performance-Alibaba-arrhythmia}, \ref{performance-Alibaba-axis}, \ref{performance-Alibaba-wave}, \ref{performance-Alibaba-others} are the classification performance from 17 labels of Alibaba dataset.
A majority of the classes display a comparable pattern to the macro-average F1 score. However, the optimal performance for each class is different from the average performance.
Certain labels, such as SA and PAC of Physionet 2021 dataset, STach and STTC of Alibaba dataset exhibit different performance trend.
This suggest that the globally optimal performance is not universally effective for all individual labels.


\begin{figure}[hb]
        \includegraphics[width=0.5\textwidth]{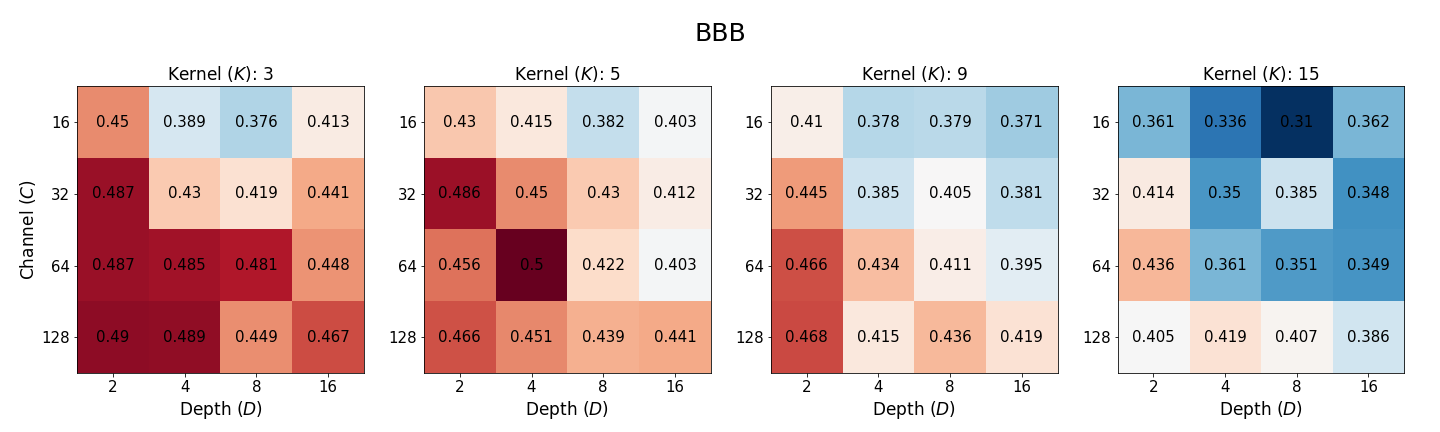}
        \includegraphics[width=0.5\textwidth]{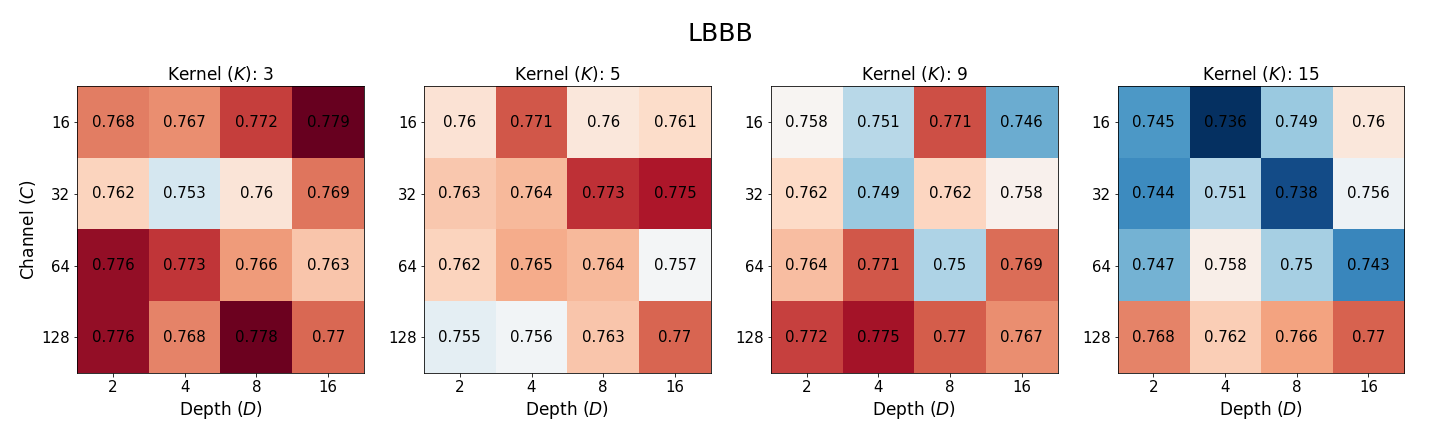}
        \includegraphics[width=0.5\textwidth]{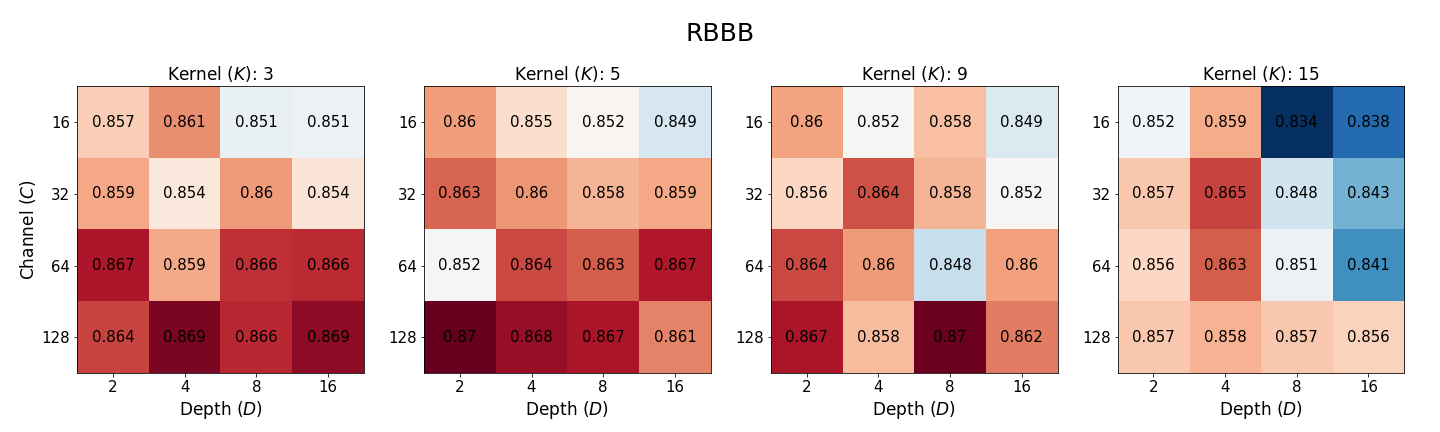}
        \includegraphics[width=0.5\textwidth]{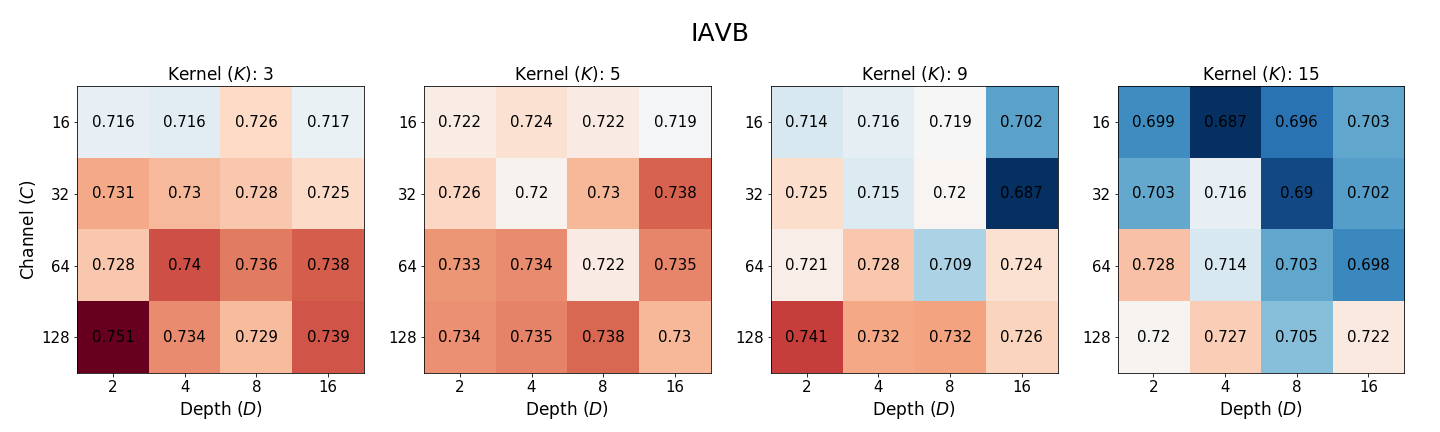}
        \includegraphics[width=0.5\textwidth]{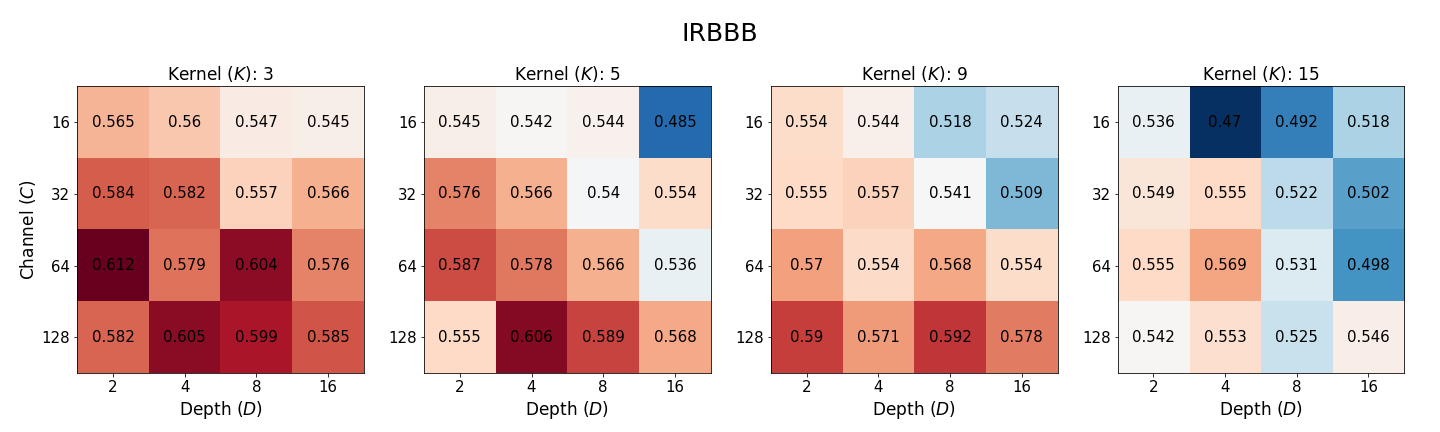}
        \includegraphics[width=0.5\textwidth]{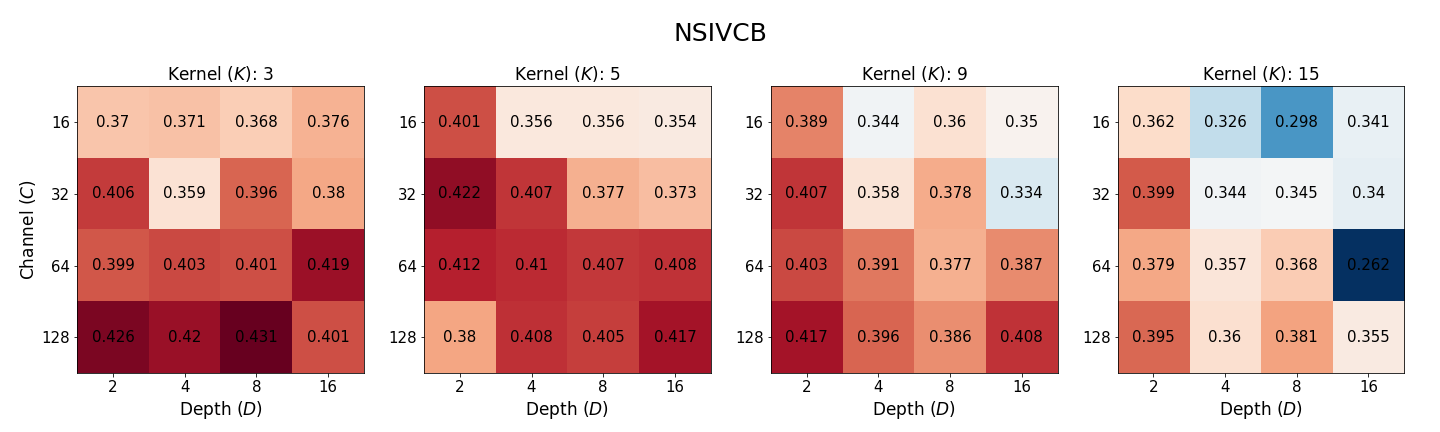}
        \includegraphics[width=0.5\textwidth]{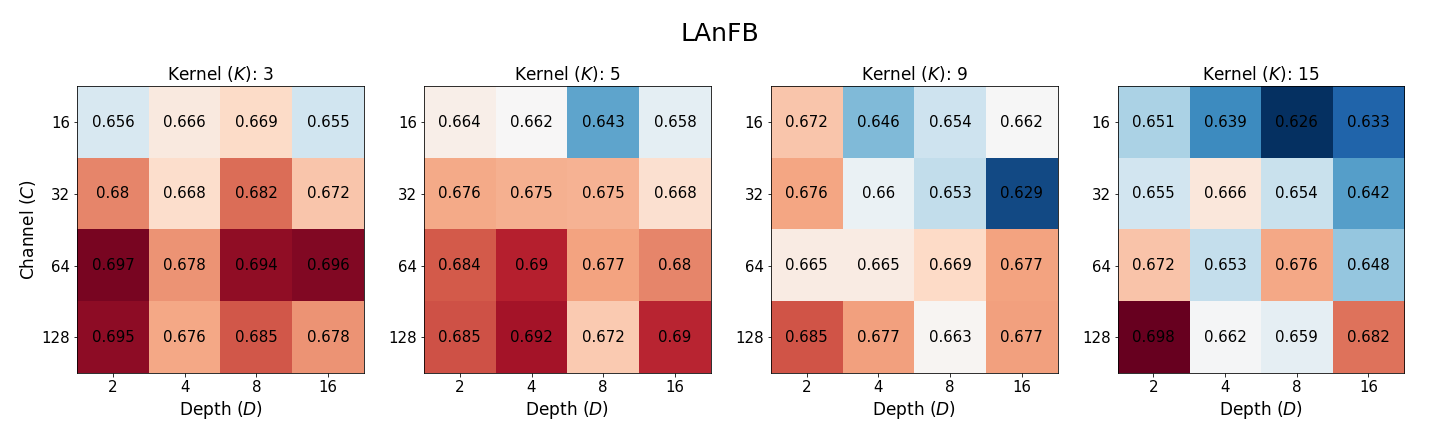}
    \caption{The classification performance of networks with different layer depth, number of channels, and kernel size on conduction disorder labels in Physionet 2021 dataset.}
    \label{performance-physionet-conduction}
\end{figure}
\vspace{-5mm}

\vspace{-20mm}

\vspace{-20mm}

\vspace{30mm}

\vspace{-15mm}

\vspace{-15mm}

\begin{figure}[h]
        \includegraphics[width=0.5\textwidth]{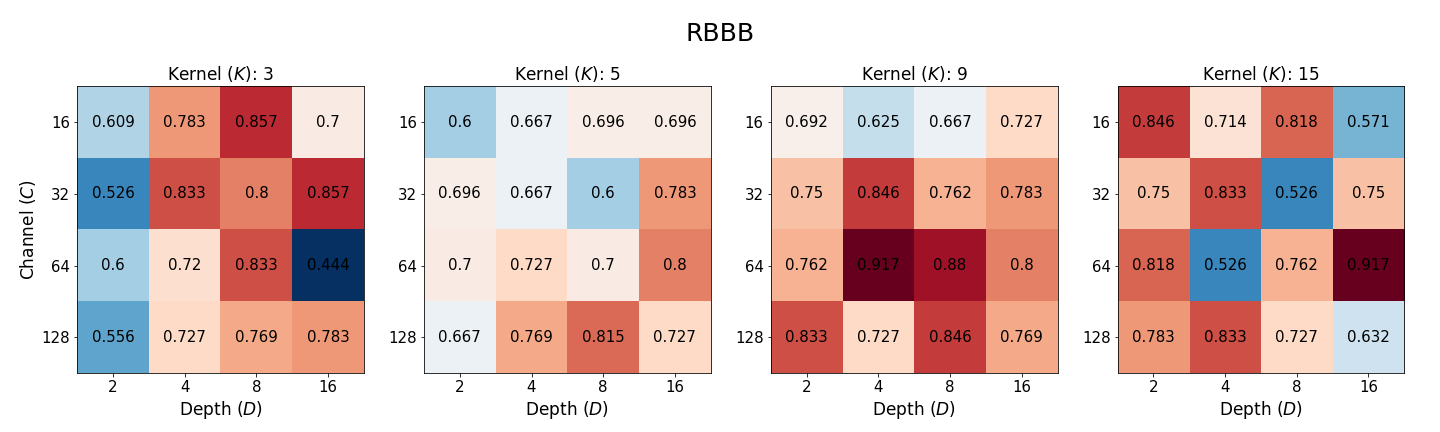}
        \includegraphics[width=0.5\textwidth]{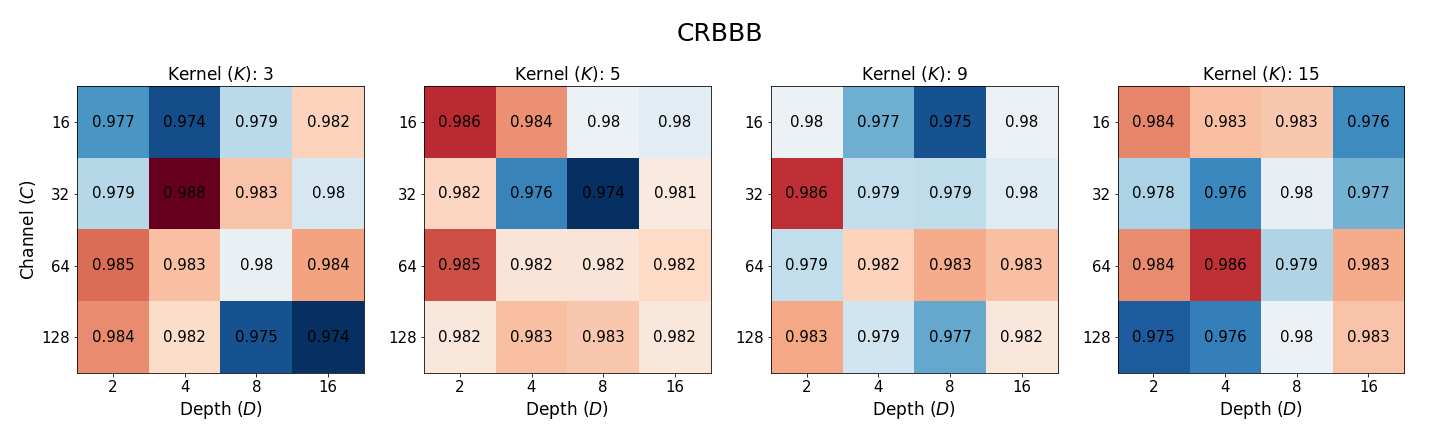}
        \includegraphics[width=0.5\textwidth]{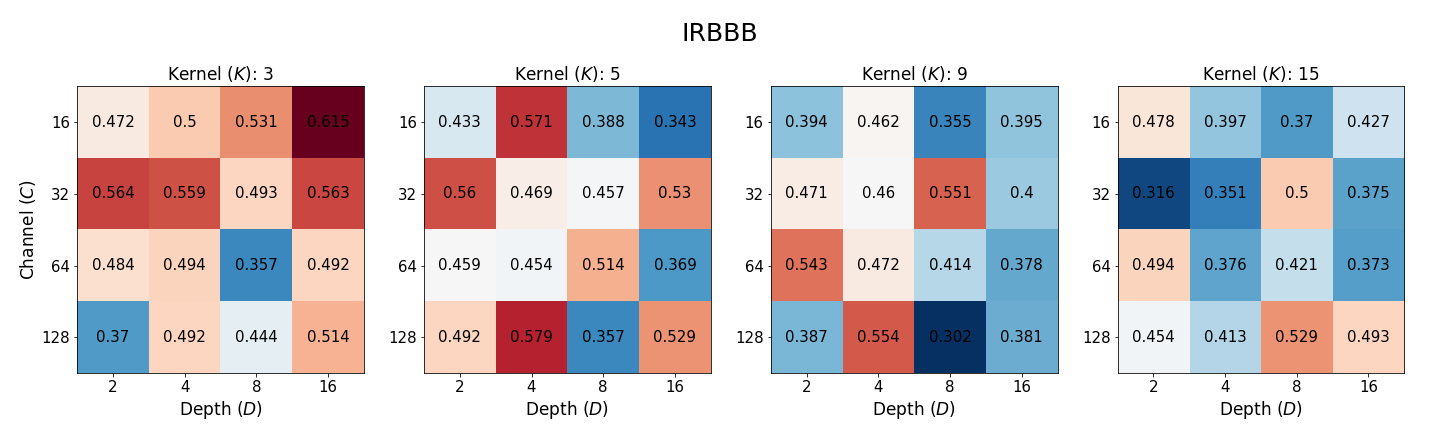}
        \includegraphics[width=0.5\textwidth]{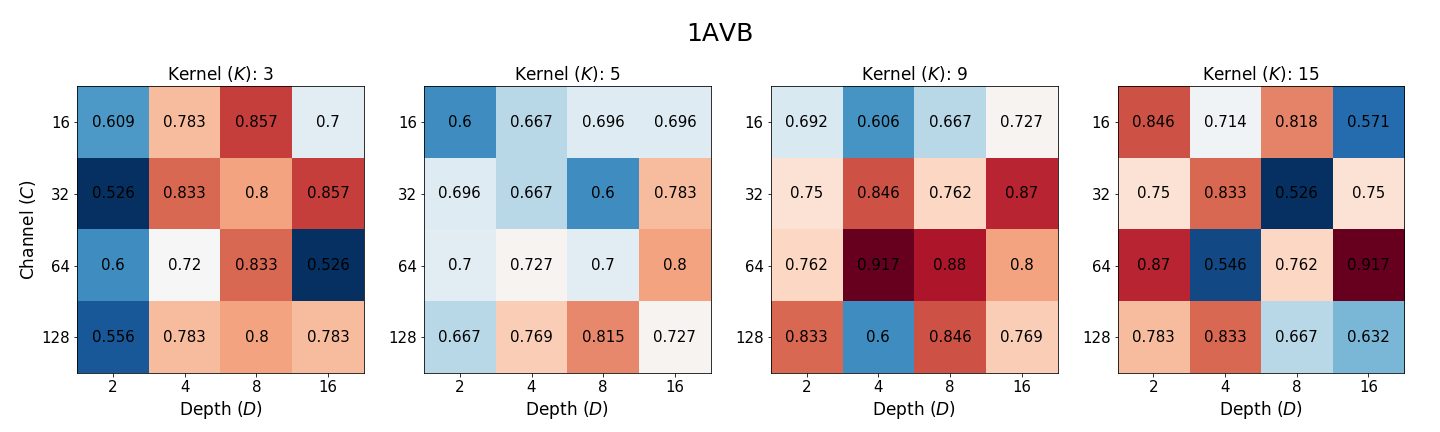}
    \caption{The classification performance of networks with different layer depth, number of channels, and kernel size on conduction disorder labels in Alibaba dataset.
    \\}
    \label{performance-Alibaba-condiction}
\end{figure}
\vspace{5mm}

\vspace{-20mm}
\vspace{-20mm}

\vspace{20mm}

\clearpage

\section{Receptive Field and Global Average Pooling}
\label{append:e}
The receptive field refers the region of input space that a particular neuron is connected. 
Generally, networks with fewer layer and small kernel size have a narrower receptive field than that with deeper layer and large kernel size. 
In our experiment, a network with small number of layer (i.e. $D$=2) and kernel size (i.e. $K$=3) spans only a few cycles. Conversely, deeper networks of those with larger kernel expand their receptive fields to cover the entire 10-second signal.
The optimal size of receptive field is different by data. A large receptive fields has benefit in extracting long-term dependencies while small receptive fields benefit the local details. In the case of ECG classification, small receptive field already include sufficient information for feature extraction. In fact, the use of a large receptive field in the ECG signals could result in the generation of redundant information, and this redundancy could introduce the challenges in the optimization process and potentially degrade the performance of the model~[\cite{zhang2020intelligent}].

Global Average Pooling is technique used in convolutional neural network for reducing spatial dimension in image or temporal dimension in time series data. 
If the receptive field is small or local, GAP serves to average the features extracted from this local patch of the input.
Consequently, it generates an average representation of the small part of the input. This can be compared to an ensemble of different view of the signal, which usually yield a generalized and robust representation.
In contrast, with a larger receptive field, the benefit of having this ensemble of varied perspectives diminishes. This is because each receptive field perceives a identical region, which can potentially reduce the advantage of ensemble of representation.

\end{document}